\documentclass{article}

\PassOptionsToPackage{numbers, compress}{natbib}
 \usepackage[preprint]{neurips_2026}

\usepackage{graphicx}       
\graphicspath{{./}}
\usepackage[utf8]{inputenc} 
\usepackage{ulem}
\usepackage{xcolor}
\usepackage[T1]{fontenc}    
\usepackage{hyperref}       
\usepackage{url}            
\usepackage{booktabs}       
\usepackage{amsfonts}       
\usepackage{amssymb}        
\usepackage{nicefrac}       
\usepackage{microtype}      
\usepackage{xcolor}         
\usepackage{tcolorbox}      
\tcbuselibrary{breakable}
\usepackage{listings}        
\usepackage{amsmath}         
\usepackage{colortbl}        
\usepackage{multirow}        
\usepackage{enumitem}        
\usepackage{float}           
\usepackage{algorithm}       
\usepackage{algpseudocode}   
\usepackage{pgfplots}        
\pgfplotsset{compat=1.18}
\usepgfplotslibrary{colorbrewer}
\usepgfplotslibrary{groupplots}
\usetikzlibrary{fit,calc}    
\usepackage[dvipsnames]{xcolor}
\usepackage{xspace}
\definecolor{pastellavender}{HTML}{EDE7F6}
\definecolor{pastelmint}{HTML}{E0F2F1}
\definecolor{pastelblue}{HTML}{E3F2FD}
\definecolor{deeppurple}{HTML}{5E35B1}
\definecolor{deepteal}{HTML}{00796B}
\definecolor{deepblue}{HTML}{1565C0}
\definecolor{lightpurpleborder}{HTML}{B39DDB}
\definecolor{lightteal}{HTML}{B2DFDB}
\definecolor{lightblueborder}{HTML}{90CAF9}
\definecolor{tableheader}{HTML}{E8EAF6}
\definecolor{maxhighlightcolor}{HTML}{EDE7F6}

\hypersetup{
    colorlinks=true,
    linkcolor=deeppurple,
    citecolor=deepteal,
    urlcolor=deepblue
}

\tcbset{
    promptbox/.style={
        colback=pastellavender,
        colframe=lightpurpleborder,
        coltitle=white,
        fonttitle=\bfseries\small,
        colbacktitle=deeppurple,
        boxrule=0.5pt,
        arc=2pt,
        left=6pt, right=6pt, top=4pt, bottom=4pt,
        fontupper=\small\ttfamily,
        breakable
    },
    mintbox/.style={
        colback=pastelmint,
        colframe=lightteal,
        coltitle=white,
        fonttitle=\bfseries\small,
        colbacktitle=deepteal,
        boxrule=0.5pt,
        arc=2pt,
        left=6pt, right=6pt, top=4pt, bottom=4pt,
        fontupper=\small\ttfamily,
        breakable
    },
    bluebox/.style={
        colback=pastelblue,
        colframe=lightblueborder,
        coltitle=white,
        fonttitle=\bfseries\small,
        colbacktitle=deepblue,
        boxrule=0.5pt,
        arc=2pt,
        left=6pt, right=6pt, top=4pt, bottom=4pt,
        fontupper=\small\ttfamily,
        breakable
    }
}
\newcommand{\shortname}{\texttt{ARTS}}
\newcommand{\fullname}{\textbf{A}gentic \textbf{R}easoning for \textbf{T}ree \textbf{S}earch}

\newif\ifcommentsoff
\commentsofftrue

\newcommand{\para}[1]{\textbf{#1}}

\tcbuselibrary{skins,breakable}

\newtcolorbox[auto counter]{finding}[1][]{%
enhanced, breakable, 
colback=white,
colframe=RoyalPurple,
boxrule=0.4pt,
arc=2pt,
left=12pt, right=10pt, top=8pt, bottom=8pt,
borderline west={2.5pt}{0pt}{RoyalPurple},
before upper={\textbf{\textcolor{RoyalPurple}{Finding~\thetcbcounter.}}\hspace{0.75em}},
#1
}

\title{ Learning the \texttt{ARTS} of Search for Automated Discovery}

\author{%
    Gurusha Juneja\thanks{University of California, Santa Barbara} \And
    Arnav Kumar Jain\thanks{Universit\'e de Montr\'eal and Mila- Quebec AI Institute} \And
    Deepak Nathani\footnotemark[1] \And
    William Yang Wang\footnotemark[1] \And
    Xin Eric Wang\footnotemark[1]
  }

\begin{document}

\maketitle

\begin{abstract}

Scientific discovery can be formulated as an iterative search process over the space of hypotheses and experiments.
Contemporary methods navigate this space using heuristics such as MCTS.
These algorithms conflate the merit of a hypothesis with the quality of its experimental execution. 
A promising hypothesis with preliminary execution is therefore ranked below a modest hypothesis whose execution is refined.
Moreover, prior methods prune the search logs as the search progresses because the accumulated history outgrows the context window. 
We propose \fullname{} (\shortname{}), where we deploy a reasoning language model to navigate this space. The model inspects prior execution logs, diagnoses whether earlier failures arose from faulty implementations or bad hypotheses, and selects the hypothesis to build on next. 
To mitigate challenges with context length, \shortname{} uses test-time training to instill the knowledge of search tree in the model weights.
Across 22 tasks from MLGym and MLEBench, we show that \shortname{} outperforms leading algorithms, with over $15.3\%$ relative improvement in the normalized score. 
With test-time training we show that a Qwen3-4B agent can match performance with closed-source frontier models like Gemini-3 Pro and GPT o3-reasoning with upto 5$\times$ lower inference cost.
We further observe that on partially observable RL tasks, the test-time trained Qwen3-4B scientist surpasses \shortname{} with the o3 scientist by rediscovering the human-best recurrent-memory solution that heuristic methods prune away.

\end{abstract}

\section{Introduction}

Scientific discovery is an iterative \textit{search process} in which multiple hypotheses are proposed, tested, and refined to arrive at novel insights.
The advent of large reasoning models have led to their use as AI Scientist  conducting this search~\citep{sakana_ai_scientist, autoresearch}.
In the past, these systems have produced new constructions in extremal combinatorics~\citep{funsearch}, improved classical algorithms for matrix multiplication~\citep{alphaevolve}, planned and executed autonomous chemical syntheses~\citep{coscientist}, and generated novel hypotheses on antimicrobial resistance~\citep{google_ai_coscientist}.  
The success of these systems often rely on large sampling budgets and repeated execution of expensive experiments ~\citep{funsearch, alphaevolve}.
This cost is significant in ML research~\citep{aira, mlgym, mlebench, rebench}, where each hypothesis must be implemented as code, trained, debugged, and evaluated.
Scaling automated research, therefore, requires efficient searching algorithms.

Existing search algorithms fall into three categories: linear search, tree-based search, and evolutionary search. 
Searching linearly ~\citep{mlgym, mlagentbench, autoresearch} makes it difficult for the agent to revisit and improve upon an initial bad hypothesis~\citep{qin2025backtrack}.
Tree-based~\citep{aira, aide, sela, imcts, mars} and evolutionary~\citep{alphaevolve, openevolve, shinkaevolve,weng2026GEA} algorithms alleviate this concern, but they search using score-based heuristics. This conflates hypothesis quality with it's execution in code. In such methods, a modest hypothesis that is well implemented is likely to be preferred over a promising hypothesis with preliminary execution. For instance, an untuned Transformer scores below a LSTM on a sequence modeling task. 
Score-dependent heuristic search would prefer exploring LSTMs further even though the Transformer might surpass them upon tuned. 
Moreover, the reasoning models driving these searches suffer from diversity collapse~\citep{verbalized_sampling} and the search procedures themselves are sensitive to the exploration-exploitation tradeoff~\citep{herr2025llmfirstsearchselfguidedexploration}, often committing to sub-optimal directions~\citep{tot_shunyu}.
Lastly, as the search progresses, the accumulated history outgrows the model's context. Existing algorithms either summarize~\citep{aira} or prune~\citep{autoresearch} this history and lose information, or retain it and reason less reliably over long contexts~\citep{liu2023lost, hsieh2024ruler, attentionsinks}.
  



\begin{figure}[t]
    \centering
    \includegraphics[width=\textwidth]{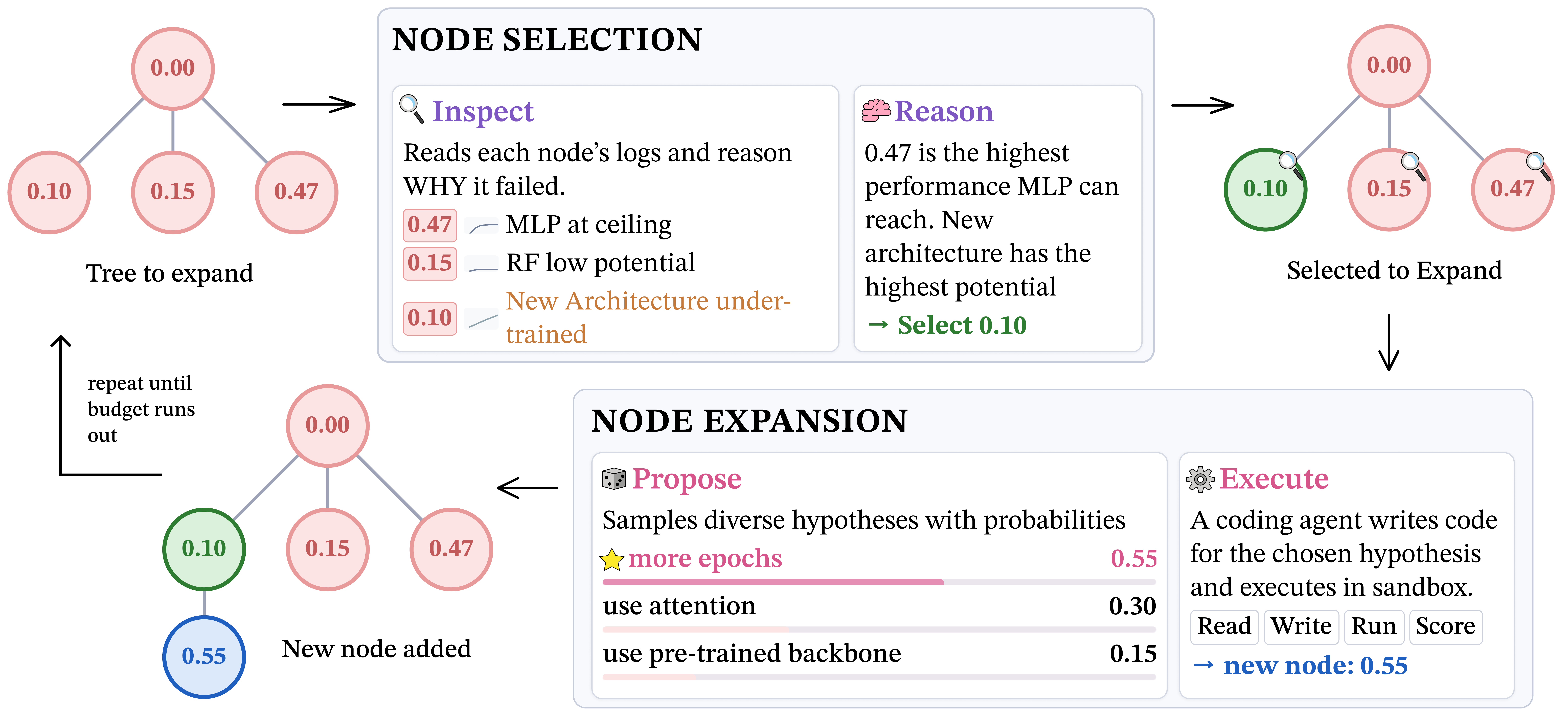}
    \caption{\textbf{A single search step of \shortname{}.} A node is one validated experiment with a hypothesis, code, logs, and score. Starting from the current search tree (\emph{Tree to expand}), the \texttt{scientist} inspects the candidate nodes and reasons about \emph{why} each scored as it did, then selects the most promising node rather than the one with the highest score. In this figure, it picks the node with score $0.10$ since the low score comes from under training rather than a weak hypothesis (\emph{Node selection}). Through verbalized sampling the \texttt{scientist} then proposes diverse hypotheses with probabilities and samples one, which the \texttt{executor} implements, runs, and scores (\emph{Node expansion}). The resulting node ($0.55$) becomes the new best and is added to the tree. This loop repeats until the budget is exhausted. \textcolor{red}{Red} nodes are explored candidates, \textcolor{Green}{green} marks the node selected for expansion, and \textcolor{blue}{blue} marks the newly added node.}
    \label{fig:llmg-stages}
    \vspace{-1.5em}
\end{figure}

In this work, we propose Agentic Reasoning for Tree Search (\shortname{})\footnote{Code and Trajectories at: \url{https://automating-discovery.github.io/arts/}.} that uses a Large Reasoning Model (LRM) to navigate the search space by exploring diverse hypotheses (Figure~\ref{fig:llmg-stages}; Algorithm~\ref{alg:llm_guided}). When the search tree exceeds the context limits, $\shortname{}^*$ can learn from prior experiments instilling the knowledge into it's weights (Algorithm~\ref{alg:ttt}).
In \shortname{}, an agentic LRM, which we call the \textit{scientist} recieves the current search tree. It can inspect any existing node to learn about the prior explored hypothesis, execution errors, and performance.
Using this information, the scientist \textit{"reasons"} to select the node for expansion and generates the next hypothesis.
This reasoning let's scientist judge if a poor score is the outcome of a fundamental limitation in the hypothesis or a correctable fault in its execution. 
To improve hypothesis diversity, \shortname{} uses an adapted version of Verbalized Sampling~\citep{verbalized_sampling}. 
Our experiments show that, \shortname{} improves the mean normalized score by $15.3\%$ compared to leading methods across tasks in MLGym and MLEBench.





$\shortname{}^*$ handles context-length overflow via test-time training~\citep{ttt}, fine-tuning the scientist using the current search tree. 
Prior work has shown that such finetuning encodes knowledge into model weights~\citep{lora_knowledge, in_weight_vs_in_context}. 
In our experiments, we find that the trained Qwen3-4B scientist reaches the performance of the o3 scientist at \textbf{5$\times$} less inference cost. 
Remarkably, $\shortname{}^*$ with Qwen3-4B scientist outperforms all other methods, including \shortname{} with o3 scientist, on MetaMaze task with an optimized LSTM-based solution. 
Heuristic methods propose the recurrent-memory hypothesis but prune it after early instability, and \shortname{} with the o3 scientist returns to it repeatedly yet drifts as context grows. Test-time training instills this search experience in the Qwen-4B scientist's weights, allowing it to return to and refine the direction.

The contributions of this paper are as follows:
\begin{itemize}[leftmargin=*]
\item We propose \textbf{Agentic Reasoning for Tree Search} (\shortname{}), in which a Large Reasoning Model inspects prior node's hypotheses, executions, and outcomes and reasons over them to drive search.
\item We propose a $\shortname{}^*$, where the scientist is \textbf{Test-Time Trained} to handle context overflow that emerges as the search history grows. The \texttt{scientist} is finetuned using the past history which help it retain information in its weights akin to a memory mechanism. 
\item On MLGym and MLEBench, \textbf{\shortname{} outperforms leading methods} on 16 of 22 tasks and improves the mean normalized score \textbf{by 15.3\%}. With $\shortname{}^*$, a Qwen3-4B scientist matches frontier closed-source models on several tasks at a fifth of the inference cost.
\end{itemize}

\section{Related Work}
\label{sec:related}

\para{ML research as search.}
Automated ML research can be viewed as search over hypotheses, code, and execution traces. Benchmarks such as MLGym~\citep{mlgym}, MLEBench~\citep{mlebench}, RE-Bench~\citep{rebench}, and DiscoveryBench~\citep{discoverybench} make this setting concrete by evaluating agents on tasks where each node is a real training run. Existing methods mainly differ in how they choose the next node to expand.

\para{Linear search.}
Linear agents follow a single trajectory and condition on their previous actions and outcomes. This includes ReAct~\citep{yao2023react}, Reflexion~\citep{shinn2023reflexionlanguageagentsverbal}, AI Scientist v1~\citep{sakana_ai_scientist}, and AutoResearch~\citep{autoresearch}. Reasoning models can revise ideas inside one generation~\citep{deepseekr1, qin2025backtrack}, but this is still bounded by the current context. In ML research, long traces of code, logs, and failed runs accumulate quickly, and long-context studies show that retrieval and reasoning degrade as context grows~\citep{liu2023lost, hsieh2024ruler}. Linear search is therefore simple, but it has weak global backtracking.

\para{Tree and evolutionary search.}
Tree-search methods such as AIDE~\citep{aide}, AIRA~\citep{aira}, and AI Scientist v2~\citep{sakana_ai_scientist} maintain explicit trees and select nodes using score-based heuristics such as greedy value or UCT. This is problematic for ML research because a node score conflates hypothesis quality with implementation quality. A good idea can fail because of a missing import, a shape mismatch, or an unstable training loop. Evolutionary systems such as FunSearch~\citep{funsearch}, AlphaEvolve~\citep{alphaevolve}, OpenEvolve~\citep{openevolve}, and MLEvolve~\citep{mlevolve} search over populations of programs, but ML training code is tightly coupled and expensive to evaluate. Mutation and crossover can be effective for compact programs, but they are a brittle primitive for full ML pipelines.

\para{Reasoning-guided search and test-time training.}
\shortname{} replaces heuristic node selection with an agentic scientist that inspects prior execution traces and chooses what to expand next. Verbalized sampling~\citep{verbalized_sampling} is used to keep the proposal distribution diverse instead of greedily following one direction. We also build on test-time training~\citep{ttt, akyurek2025the, ttt-discover, si2026execution}, but train a tree-structured scientist rather than a one-turn proposal policy. The extended discussion is in Appendix~\ref{app:extended_related_work}.

\section{Research as a Search Problem}
\label{sec:background}

\para{Research tasks.}
For a Machine Learning research task, we provide the agent with an initial workspace $\mathcal{W}_0$, an evaluation metric $m$, a baseline score $s_0$, and a wall-clock budget $T$.
We denote the task by $\mathcal{P}=(\mathcal{W}_0,m,s_0,T)$.
The workspace contains task instructions, data access, and an evaluation harness.
The agent interacts with the workspace by editing code inside a sandbox and querying the evaluator.
The evaluator returns execution logs and a score when the run succeeds.
When evaluation fails, it returns error messages and partial logs.

\para{Experiments and search histories.}
Each evaluated experiment creates a node $v$ in the search tree $\mathcal{G}_t=(V_t,E_t)$.
The node stores a parent $p_v$, a hypothesis $h_v$, a code state $c_v$, execution logs $\ell_v$, a score $s_v$, and a depth $d_v$.
The parent's code state $c_{p_v}$ serves as the starting point on which the agent builds the new code implementation $c_v$ for $h_t$.
The search history $\mathcal{G}_t$ can be a chain, a tree, or a graph.

At step $t$, the research agent chooses an action $a_t=(p_t,h_t,c_t)$, where $p_t\in V_t$ is the parent node to expand, $h_t$ is the hypothesis to test, and $c_t$ is the code implementation for $h_t$ starting from the parent code state $c_{p_t}$. 
The evaluator runs $c_t$ inside the sandbox and returns logs $\ell_t$ and score $s_t$.
We call the component that proposes $h_t$ the \texttt{scientist}.
We call the component that writes $c_t$ the \texttt{executor}.
Prior methods~\citep{aira,mlevolve,autoresearch} use the same agent as both \texttt{scientist} and \texttt{executor}, and pick the parent either as the last node (Linear) or by a score heuristic like UCB (MCTS, evolutionary).

\section{Agentic Reasoning for Tree Search}

We present \shortname{} (\textbf{A}gentic \textbf{R}easoning for \textbf{T}ree \textbf{S}earch) that leverages two design choices.
First, we use different models for the scientist and the executor, since choosing a research direction and writing correct code require different abilities.
Second, instead of selecting the parent node $p_t$ with a fixed heuristic, we let the scientist choose both the parent $p_t$ and the hypothesis $h_t$.
The scientist is shown a compact view of the tree, including each node's identifier, score, and hypothesis.
It also has access to persistent memory, and can inspect prior nodes to read their code and execution logs.
To mitigate diversity collapse, we use verbalized sampling~\citep{verbalized_sampling}.
The scientist outputs a proposal distribution over $K$ candidate actions, and the system samples from it rather than always taking the top-ranked action.
The executor then implements the sampled hypothesis from the selected parent's code state.

\subsection{The \texttt{Scientist}}
\label{subsec:scientist}
The \texttt{scientist} selects which node to expand and what hypothesis to try, both requiring reasoning over multiple past experiments. 
For instance, concluding that the bottleneck is the loss function and not the model capacity after finding that (a) increasing model capacity failed to improve validation accuracy, and (b) training curves show that training accuracy also failed to improve, requires strong reasoning capabilities.
We therefore use o3 as the \texttt{scientist} owing to it's strong multi-step reasoning abilities~\citep{openai_2025, chollet2024o3arc, glazer2024frontiermath, rein2024gpqa}. Below, we describe the components used in Algorithm~\ref{alg:llm_guided}.

\para{Node Inspection.} A node's score alone does not give enough information to judge why it succeeded or failed. A node can have low score due to a number of reasons, including wrong hypothesis, wrong code execution, correct hypothesis but insufficient training time, etc. Without knowing the exact reason for failure or success, the next hypothesis is just a guess. 
The $\textsc{Inspect}$ tool returns the code, training curves, and program output of any prior node in the search tree. This lets the \texttt{scientist} identify the actual cause, for example, observing that a sparse-reward run never received a positive reward, and proposing reward shaping rather than an bigger model. After $\textsc{Inspect}$, the \texttt{scientist} reasons to choose the appropriate next node to expand and the hypothesis to test.

\para{Baseline Audit.} Many tasks have structural biases that can be exploited to arrive to clever solutions. For example, on the Vesuvius ink-detection task only a handful of samples have labels, but the unlabeled data is huge; self-supervised pretraining can be exploited in such cases. To ensure that the \texttt{scientist} is aware of these biases while reasoning about hypotheses, we reserve the first $R$ calls to the \texttt{scientist} as \textsc{audits}. During this phase no node may be expanded and no code may be changed; the \texttt{scientist} may only read the data, baseline code, and evaluation script. The audit must produce statements about the task, such as the structure of the metric (e.g.\ ordinal, sparse, class-imbalanced) or the largest information-loss step in the baseline.

\para{Hypothesis sampling.} Exploring diverse solutions is essential in research. If we ask the scientist to provide the next hypothesis to explore, it returns most probable continuation, which usually are small tweaks over the learning rate. To ensure that the explored solution space is diverse, we use verbalised sampling~\citep{verbalized_sampling}. Here, the scientist enumerates $K$ candidate hypothesis $h_t^k$ each associated with a probability $\pi_t^k$ assigned by the scientist to the hypothesis. One candidate is drawn by an external sampler, since when the scientist samples from its own list it tends to pick the first candidate. At deeper nodes we ask the scientist to sample from the tail of the hypothesis distribution.

\para{Memory.} Over a long search the tree accumulates hundreds of nodes, and re-reading them all on every step would exhaust the context window. We complement the scientist with an editable memory module. After each expansion it writes one short insight (the score achieved, the failure mode, etc.) to this memory, appended only if not already present. The scientist can read this memory at all times, including when selecting a node and generating the next hypothesis.

\subsection{The \texttt{Executor}}

The \texttt{executor} is a coding agent that takes the \texttt{scientist's} hypothesis $h_t$ and the selected parent's validated code $c_{p_t}$, and writes a new code state $c_t$. We use Gemini 3 Flash as the \texttt{executor} since it is fast, inexpensive, and strong at code generation~\citep{gemini3}. The executor has access to three tools: \textsc{read\_file} (to read existing code), $\textsc{write\_file}$ (to write new code), and $\textsc{validate}$ (to execute the code in the sandbox and get the validation score). It can edit and validate the code until either a valid score is returned or the maximum action
budget is reached. It is prompted to implement $h_t$ as given by the scientist, we find that without this rule it substitutes $h_t$ with alternatives of its own. Once the executor produces a successful $\textsc{validate}$, the resulting node is appended to the search tree with its score $s_t$ (or null if no valid submission was produced within the action budget).

\begin{figure}[t]
\centering
\footnotesize
\newlength{\algorithmblockheight}
\setlength{\algorithmblockheight}{0.25\textheight}
\begin{minipage}[t][\algorithmblockheight][t]{0.49\linewidth}
\hrule
\vspace{0.35em}
\refstepcounter{algorithm}
\label{alg:llm_guided}
\textbf{Algorithm~\thealgorithm: \shortname{} search.}
\vspace{0.35em}
\hrule
\vspace{0.35em}

\begin{algorithmic}[1]
\State \textbf{Input:} Task $\mathcal{P}$, time budget $T$, \texttt{scientist}, \texttt{executor}
\State \textbf{Initialize:} Tree $\mathcal{G}$ = (code $c_{\text{base}}$, score $s_{\text{base}}$)
\State \textbf{\textsc{Audit:}} data, code, and metric. \textcolor{deepteal}{// No node expanded.}
\While{time < T}
    \State \textcolor{deepteal}{// scientist calls inspects tree to get logs}
    \State $\ell_t \leftarrow \textsc{Inspect}(\mathcal{G}_t, \mathcal{P})$ 
    \State \textcolor{deepteal}{//parent selection and hypothesis sampling $(h_t^k,\pi_t^k)$}
    \State $p_t,\{(h_t^k,\pi_t^k)\}_{k=1}^{K} \leftarrow \texttt{scientist}(\ell_t, \text{memory})$
    \State \textcolor{deepteal}{// external sampler draws one hypothesis}
    \State $h'_t \leftarrow \textsc{Sample}(\{(h_t^k,\pi_t^k)\}_{k=1}^{K})$
    \State \textcolor{deepteal}{// implement and validate}
    \State $c_t, s_t \leftarrow \textsc{Execute}(p_t,h'_t, c_{p_t})$ 
    \State $\mathcal{G}_{t+1} \leftarrow \mathcal{G}_t \cup (p_t,h'_t,c_t, s_t)$ \textcolor{deepteal}{// add new node to tree}
\EndWhile
\State \Return best validated node
\end{algorithmic}
\vfill
\hrule
\end{minipage}
\hfill
\begin{minipage}[t][\algorithmblockheight][t]{0.49\linewidth}
\hrule
\vspace{0.35em}
\refstepcounter{algorithm}
\label{alg:ttt}
\textbf{Algorithm~\thealgorithm: $\shortname{}^*$ with test time training.}
\vspace{0.35em}
\hrule
\vspace{0.35em}

\begin{algorithmic}[1]
\State \textbf{Input:} Task $\mathcal{P}$, group size $B$, \texttt{scientist} $\pi_\theta$, \texttt{executor}
\State \textbf{Initialize:}  $\mathcal{G}$ = ( $c_{\text{base}}$,  $s_{\text{base}}$) LoRA adapters. 
\For{each GRPO step}
    \State $\ell_t \leftarrow \textsc{Inspect}(\mathcal{G}_t, \mathcal{P})$ 
    \State \textcolor{deepteal}{// parent selection and hypothesis sampling}
    \State $\{(p_i,\{(h_i^k,\pi_i^k)\}_{k=1}^{K})\}_{i=1}^{B} \sim \pi_\theta(\mathcal{G}_t)$
    \State $h'_i \leftarrow \textsc{Sample}(\{(h_i^k,\pi_i^k)\}_{k=1}^{K})$  for each $i$
    \State $c_i, s_i \leftarrow \texttt{executor}(p_i,h_i, c_{p_t})$ for each $i$
    \State \textcolor{deepteal}{// percentile reward}
    \State $r_i \leftarrow R(s_i,\text{tree})$ 
    \State Update LoRA with GRPO on $\{(p_i,h_i,r_i)\}_{i=1}^{B}$
    \State \textcolor{deepteal}{// Add rollouts to the tree.}
    \State  $\mathcal{G}_{t+1} \leftarrow \mathcal{G}_t \cup \{(p_i,h_i,c_i, s_i)\}_{i=1}^{B}$
\EndFor
\State \Return trained scientist $\pi_{\theta'}$, best validated node
\end{algorithmic}
\vfill
\hrule
\end{minipage}
\end{figure}

\subsection{$\shortname{}^*$: Test-Time Training the \texttt{Scientist}}
\label{subsec:ttt}

The search tree $\mathcal{G}_t$, especially on hard problems like drug discovery, can grow exponentially with depth and exceed \texttt{scientist}'s context limits. A tree with depth 8 and branching factor 3 has over 6000 nodes. Prior algorithms prune earlier nodes in the search tree.  Further, it is also known that long contexts degrade reasoning quality of models~\citep{liu2023lost, attentionsinks}. To solve this problem we propose test time train (TTT) the \texttt{scientist} on the search tree it has explored so far. We fine-tune the model with LoRA~\citep{hu2022lora} at regular intervals to distill the information from experiments in the model weights. Algorithm~\ref{alg:ttt} summarizes the $\shortname{}^*$ procedure. 

\para{Rollout.} Prior work~\citep{ttt-discover} has explored TTT for hypothesis generation, but the action is only the generated hypothesis and node selection is heuristic. In our setting we make two decisions per step, which node $p_t$ in the search tree to expand from, and what hypothesis $h_t$ to test. Given the state $\mathcal{G}_t$ (see \S\ref{sec:background}), the \texttt{scientist} agent outputs the action $a_t = (p_t, h_t)$. We sample $N$ such actions from the current state $\mathcal{G}_t$ and compute a reward for each, as explained below. The $N$ rewards form a GRPO group and is used to update the adapter weights. Since each action creates a new node, they are appended to the tree before the next batch of rollouts is sampled.
This teaches the model both which parent to pick and what hypothesis to explore for the largest gains in the validation score.

\para{Reward.} The reward is based on the score $s_t$ that the action $a_t = (p_t, h_t)$ receives after the code implementation $c_t$ by the \texttt{executor} is executed. We find that single-step percentile-based rewards work best:
\begin{equation}
r_t = \begin{cases}
-0.5 & s_t = \text{null},\\
-0.2 & s_t < s_\text{base},\\
0 & s_\text{base} \leq s_t < s_{70}(\mathcal{G}_t),\\
1 & s_t \geq s_{70}(\mathcal{G}_t),
\end{cases}
\end{equation}
 where $s_{70}(\mathcal{G}_t)$ is the 70th percentile of all scores observed up to timestep $t$. This signal is denser than rewarding only the single best node, since the top 30\% of nodes get a positive reward than just the maximum one. The percentile based adaptive threshold rises as the policy improves, preventing reward saturation. Learning with one-step rewards~\cite{jain2025multiturn} works better than rewarding longer multi-step trajectories because it can provide clearer credit assignment to each action.

\para{Diversity.} Training on score based rewards can lead to diversity collapse, biasing the policy towards solutions that gave higher results early. But, we find that this is not the case even after training, since verbalized sampling ensures that the samples hypotheses in one group are diverse (See $\S$\ref{subsec:ablations}). 
\begin{table}[t]
  \caption{Performance comparison of \shortname{} with prior works and human best scores. We test on 22 tasks across MLGym and MLEBench (3 easy, 3 medium, and 4 hard tasks). The table reports the mean and standard error of the best scores achieved in every run. Every experiment is ran 3 times. Human Best is Kaggle top-1 for MLEBench and published SOTA for MLGym. Highlighted cells show the best performing methods.}
  \label{tab:main}
  \centering
  \setlength{\tabcolsep}{4.8pt}
  \renewcommand{\arraystretch}{1.05}
  \footnotesize
  \begin{tabular}{l c ccc c c}
    \toprule
    & & \multicolumn{3}{c}{\textit{Prior Works}} & \multicolumn{1}{c}{\textit{Ours}} & \\
    \cmidrule(lr){3-5} \cmidrule(lr){6-6}
    \textbf{Task} & \textbf{Baseline} & \textbf{Linear} & \textbf{AIRA} & \textbf{MLEvolve} & \textbf{\shortname{}} & \textbf{Human Best} \\
    \midrule
    \rowcolor{pastellavender}
    \multicolumn{7}{c}{\textbf{\textit{\textcolor{deeppurple}{MLGym}}}} \\
    Titanic {\scriptsize(acc $\uparrow$)} & 0.766 & 0.951$_{\pm.004}$ & 0.944$_{\pm.001}$ & 0.946$_{\pm.002}$ & \cellcolor{maxhighlightcolor}\textbf{0.984}$_{\pm.004}$ & 0.830 \\
        CIFAR-10 {\scriptsize(acc $\uparrow$)} & 0.497 & \textbf{0.956}$_{\pm.002}$ & \textbf{0.964}$_{\pm.002}$ & \textbf{0.959}$_{\pm.003}$ & \cellcolor{maxhighlightcolor}\textbf{0.971}$_{\pm.017}$ & 0.994 \\
    Fashion MNIST {\scriptsize(acc $\uparrow$)} & 0.848 & 0.946$_{\pm.006}$ & 0.947$_{\pm.000}$ & 0.948$_{\pm.001}$ & \cellcolor{maxhighlightcolor}\textbf{0.958}$_{\pm.001}$ & 0.968 \\
        House Price {\scriptsize($R^2$ $\uparrow$)} & 0.880 & 0.940$_{\pm.000}$ & \cellcolor{maxhighlightcolor}\textbf{0.944}$_{\pm.001}$ & \textbf{0.943}$_{\pm.001}$ & 0.939$_{\pm.003}$ & 0.990 \\
    \midrule

    MNLI {\scriptsize(acc $\uparrow$)} & 52.51 & \textbf{84.42}$_{\pm0.05}$ & 83.77$_{\pm0.05}$ & \textbf{84.26}$_{\pm0.08}$ & \cellcolor{maxhighlightcolor}\textbf{84.71}$_{\pm0.49}$ & 92.50 \\
    Lang.\ Modeling {\scriptsize(loss $\downarrow$)} & 4.673 & 3.986$_{\pm.169}$ & 4.673$_{\pm.000}$ & 4.015$_{\pm.146}$ & \cellcolor{maxhighlightcolor}\textbf{3.827}$_{\pm.088}$ & 3.500 \\
    \midrule
    Battle of Sexes {\scriptsize(payoff $\uparrow$)} & 1.023 & 1.448$_{\pm.001}$ & 1.442$_{\pm.001}$ & 1.446$_{\pm.000}$ & \cellcolor{maxhighlightcolor}\textbf{2.000}$_{\pm.000}$ & 1.667 \\
    Prisoner's Dilem.\ {\scriptsize(payoff $\uparrow$)} & 2.372 & 2.453$_{\pm.102}$ & 2.501$_{\pm.129}$ & \textbf{2.635}$_{\pm.012}$ & \cellcolor{maxhighlightcolor}\textbf{2.857}$_{\pm.225}$ & 3.000 \\
    Blotto {\scriptsize(score $\uparrow$)} & $-$0.248 & $-$0.076$_{\pm.327}$ & 0.247$_{\pm.003}$ & \cellcolor{maxhighlightcolor}\textbf{0.250}$_{\pm.001}$ & \textbf{0.249}$_{\pm.001}$ & 0.500 \\
    \midrule

    MountainCar {\scriptsize(reward $\uparrow$)} & 33.79 & 45.44$_{\pm6.27}$ & 80.82$_{\pm11.75}$ & 84.84$_{\pm6.38}$ & \cellcolor{maxhighlightcolor}\textbf{95.73}$_{\pm.69}$ & 99.00 \\
    Breakout {\scriptsize(reward $\uparrow$)} & 48.82 & 64.03$_{\pm10.75}$ & 57.94$_{\pm3.37}$ & \cellcolor{maxhighlightcolor}\textbf{83.28}$_{\pm3.84}$ & 78.00$_{\pm2.23}$ & 100.00 \\
    Meta Maze {\scriptsize(reward $\uparrow$)} & 15.73 & 46.80 $_{\pm2.61}$ & 36.42$_{\pm 6.47}$ & 45.35 $_{\pm3.27  }$ & \cellcolor{maxhighlightcolor}\textbf{51.20}$_{\pm1.03}$ & 52.50 \\
    \midrule
    \rowcolor{pastelmint}
    \multicolumn{7}{c}{\textbf{\textit{\textcolor{deepteal}{MLEBench}}}} \\
    Spaceship Titanic {\scriptsize(acc $\uparrow$)} & 0.000 & 0.825$_{\pm.006}$ & 0.831 & 0.834$_{\pm.001}$ & \cellcolor{maxhighlightcolor}\textbf{0.836}$_{\pm.001}$ & 0.828 \\
    Nomad 2018 {\scriptsize(MCW-RMSLE $\downarrow$)} & 1.000 & \textbf{0.063}$_{\pm.001}$ & \cellcolor{maxhighlightcolor}\textbf{0.062}$_{\pm.001}$ & \textbf{0.063}$_{\pm.000}$ & 0.064$_{\pm.001}$ & 0.051 \\
    Jigsaw Toxic {\scriptsize(CW-AUC $\uparrow$)} & 0.500 & \textbf{0.980}$_{\pm.000}$ & \textbf{0.980}$_{\pm.000}$ & \textbf{0.980}$_{\pm.000}$ & \cellcolor{maxhighlightcolor}\textbf{0.980}$_{\pm.000}$ & 0.989 \\
    \midrule
    APTOS 2019 {\scriptsize(QWK $\uparrow$)} & 0.000 & \textbf{0.926}$_{\pm.001}$ & 0.922$_{\pm.006}$ & 0.914$_{\pm.004}$ & \cellcolor{maxhighlightcolor}\textbf{0.930}$_{\pm.004}$ & 0.936 \\
    Plant Pathology {\scriptsize(MCW-AUC $\uparrow$)} & 0.500 & 0.994$_{\pm.001}$ & 0.997$_{\pm.000}$ & \cellcolor{maxhighlightcolor}\textbf{0.998}$_{\pm.000}$ & 0.995$_{\pm.003}$ & 0.984 \\
    Histopath.\ Cancer {\scriptsize(AUC $\uparrow$)} & 0.500 & 0.990$_{\pm.001}$ & \textbf{0.995}$_{\pm.000}$ & 0.994$_{\pm.000}$ & \cellcolor{maxhighlightcolor}\textbf{0.995}$_{\pm.000}$ & 1.000 \\
    \midrule
    Vesuvius Ink Det.\ {\scriptsize($F_{0.5}$ $\uparrow$)} & 0.000 & 0.479$_{\pm.067}$ & 0.309$_{\pm.112}$ & \textbf{0.549}$_{\pm.011}$ & \cellcolor{maxhighlightcolor}\textbf{0.551}$_{\pm.021}$ & 0.831 \\
    Kuzushiji Recog.\ {\scriptsize(F1 $\uparrow$)} & 0.000 & 0.894$_{\pm.040}$ & 0.872$_{\pm.026}$ & \cellcolor{maxhighlightcolor}\textbf{0.921}$_{\pm.009}$ & 0.843$_{\pm.034}$ & 0.950 \\
    HMS Brain Activity {\scriptsize(KL-div $\downarrow$)} & 1.462 & 0.543$_{\pm.055}$ & 0.550$_{\pm.020}$ & 0.583$_{\pm.013}$ & \cellcolor{maxhighlightcolor}\textbf{0.499}$_{\pm.008}$ & 0.272 \\
    RSNA Brain Tumor {\scriptsize(AUC $\uparrow$)} & 0.500 & 0.638$_{\pm.005}$ & 0.649$_{\pm.014}$ & \textbf{0.656}$_{\pm.011}$ & \cellcolor{maxhighlightcolor}\textbf{0.673}$_{\pm.021}$ & 0.621 \\
    \bottomrule
  \end{tabular}
\end{table}

\section{Experiments}
\label{sec:experiments}


\para{Tasks.} We evaluate on 22 tasks from MLGym~\citep{mlgym} and MLEBench~\citep{mlebench} \footnote{Not the entire suite of tasks because of compute and API constraints}. MLGym contains regression, Language Modeling, Vision, RL and Game Theory tasks. MLEBench consists of tasks taken from kaggle competitions, and we evaluate on 4 hard, 3 medium and 3 easy tasks. These benchmarks have well-defined metrics and realistic ML tasks. Each task provides the initial workspace $\mathcal{W}_0$ (see $\S$~\ref{sec:background}) and a held-out test set. We report human-best scores using Kaggle top-1 for MLEBench and achieve SOTA on MLGym. See Appendix~\ref{app:references_to_human_best} for the sources used. 

\para{Baselines.} We compare \shortname{} with three prior search methods from $\S$~\ref{sec:related}. Linear Search uses the AutoResearch prompt~\citep{autoresearch}, Tree Search uses AIRA (MCTS)~\citep{aira}, and Evolutionary Search uses MLEvolve~\cite{mlevolve}, and methods receive same executor, container, action set, and runtime budgets. 

\para{Models.} We use OpenAI o3~\citep{openai_2025} as the scientist and Gemini 3 Flash~\citep{gemini3} as the executor. 
The prompts for both the scientist and executor are in Appendix~\ref{app:scientist_prompts}. We also experiment with Qwen3-4B-Instruct~\citep{qwen3} as the scientist, and update it with test-time training to handle long context.

\para{Evaluation.} All methods run inside Apptainer containers with the same action set, executor sandbox, and 8-hour wall-clock budget. 
Akin to prior approaches that evaluate with fixed-budget runs~\citep{autoresearch, aira, mlevolve, mlebench}, we use time rather than node count in our experiments. We limit the runtime budget to 8 hours. Since agents sometimes can modify evaluation scripts, we restore them separately for validation (see Appendix~\ref{app:experimental_details}). 
During evaluation, we evaluate over 3 independent runs and report the mean and standard error of the best validation score reached in a run.
Each method in Table~\ref{tab:main} is given one 40\,GB A100 GPU to train any downstream model during the search process. We provide details about implementation, prompts, finetuning with RL, and compute in Appendix~\ref{app:experimental_details}.
To compute the normalized score for rliable~\cite{agarwal2021deep} we use the baseline and the human-best score. 
 

\section{Results}

Through our experiments, we answer four questions. First, does \shortname{} search more effectively than prior methods? It reaches the best score on $16$ of $22$ tasks under the same budget (\S\ref{subsec:arts-result}). Second, 
what makes \shortname{} better? We find that these gains come from three properties of its search: failure attribution, the diversity of the hypotheses it explores, and their quality~\S\ref{subsec:arts-reasoning} Third, does test-time training help? It improves a Qwen3-4B scientist to match the o3 scientist on half the tasks at a fraction of the inference cost ~\S\ref{subsec:ttt-qualitative} and finally, ablations isolate the contribution of each components in ~\S\ref{subsec:ablations}.

\subsection{How does \shortname{} compare to prior methods?}
\label{subsec:arts-result}

Table~\ref{tab:main} reports per-task performance. \shortname{} has largest gains on the harder tasks from MLEBench and MLGym where each hypothesis is an expensive training run. There is no obvious text-book obvious recipe to solve these tasks and committing to a poor choice early leads to sub-optimal utilization of the search budget. Solving them requires reasoning over what to try next, and which hypothesis has the most potential. \shortname{} performs best on most of the hardest tasks, including HMS Brain Activity and RSNA Brain Tumor, where it even beats the human best. Problems like House Price and Jigsaw Toxic are close to saturated, every method is already near the ceiling.

\begin{figure}[t]
    \centering
    \vspace{-0.6em}
    \includegraphics[width=\textwidth]{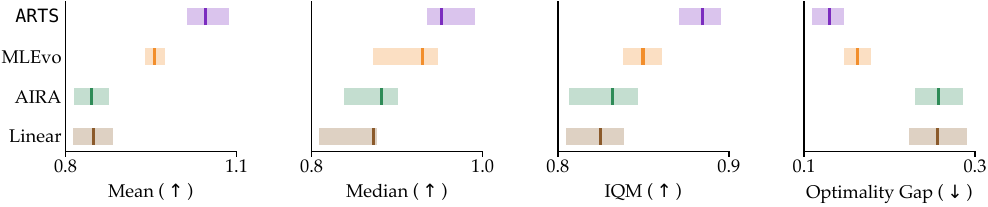}
    \vspace{-2.0em}
    \caption{\textbf{Aggregate normalized performance}. 
    We report the reliable metrics~\cite{agarwal2021deep} and observe that \texttt{ARTS} significantly  outperforms baselines at Mean and IQM while achieving lower optimality gap.
    }
    \label{fig:statistical_summaries}
\end{figure}

Figure~\ref{fig:statistical_summaries} aggregates the per-task scores into the reliable metrics of~\citet{agarwal2021deep}. \shortname{} attains the highest inter-quartile mean (IQM) among all the contemporary methods, this shows that the typical performance of \shortname{} is better than prior methods. Because IQM trims the top and bottom $25\%$ of runs, this score cannot be inflated by a few lucky outliers. The optimality gap measures how far a method stays below the human best score reached on each task. We find that \shortname{} closes most of the distance to the per-task optimum. These results show that improvement is reliable and not tail driven.

\begin{figure}[t]
    \centering
    \includegraphics[width=\textwidth]{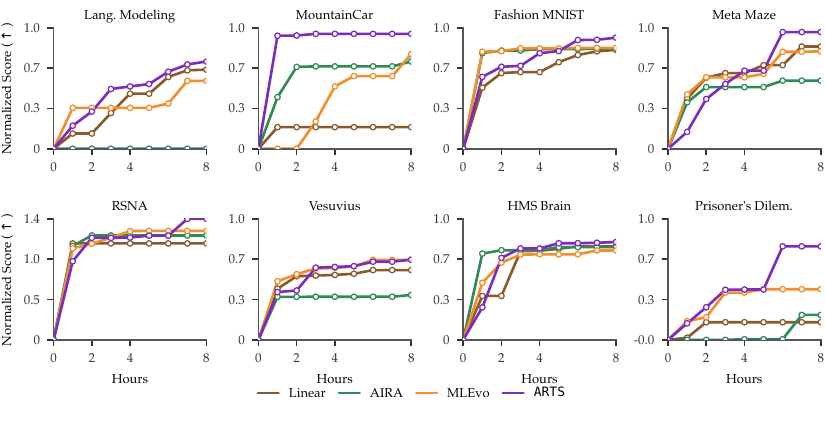}
    \vspace{-2.4em}
    \caption{Hourly progression of normalized score on randomly selected tasks under the same 8-hour budget. Lines show mean(of the best) performance across 3 runs at each hour. All baselines use same executor agent (Gemini-3-Flash) as \shortname{}.}
    \label{fig:hourlyI}
\end{figure}

Figure~\ref{fig:hourlyI} plots the best score against wall-clock time under the same $8$-hour budget. We observe that \shortname{} typically trails for the first one or two hours, catches up mid-run, and finishes above the strongest contemporary method. This pattern of a slow start followed by a higher ceiling shows that \shortname{} spends its early budget exploring broadly instead of committing onto the first locally good hypothesis, as opposed to other methods,. leading to better utilizaton of the inference budget.



\subsection{Why does \shortname{} outperform prior methods?}
\label{subsec:arts-reasoning}
We trace \shortname{}'s gains to three factors: \textit{failure attribution}, the \textit{diversity} of hypotheses it explores, and the \textit{quality} of each hypothesis, as described below.

\begin{finding}
\shortname{} does not \textbf{prematurely abandon a promising hypothesis.}
\end{finding}
Prior tree-search methods do not separate why a node scored low from
the fact that it did. A logical coding bug, bad hyperparameters, an
under-trained model, or a genuinely poor hypothesis all surface as the
same low score. The next draft pivots to a different hypothesis
even when the original one was sound. \shortname{} is designed to explicitly enforce this 
distinction, the \texttt{scientist} inspects nodes's code and reasons whether the idea was wrong or the implementation was. If the implementation
was wrong, the next experiment stays on the same axis as the
chosen node's hypothesis with corrected instructions for the executor; if
the idea was wrong, the search is free to pivot to a different
axis. This ensures that a promising idea is no longer discarded prematurely.

For instance, on HMS Brain Activity task (Fig.~\ref{fig:hms_tree}; lower KL is better; baseline = 1.462). \shortname{} drafts a pretrained ResNet-50 over 3-channel log-mel spectrograms at depth 1 and reaches \textbf{KL 0.557}. The next two children, a backbone swap to EfficientNet-B0 (0.702) and an added SpecAugment (0.565), both regress.
A score-only selection algorithm, AIRA for instance, reads these regressions as evidence that the spectrogram-CNN family is exhausted, and its next two drafts pivot to hand-crafted
features, and the children of those pivots diverge catastrophically to KL 4.08 and 9.43 while the original spectrogram-CNN family is never refined. \textcolor{blue}{\shortname{} instead re-reads the two regressed nodes' training logs, observes that the validation KL is still decreasing at the last epoch in both, and
classifies them as implementation-wrong} (under-trained under heavy augmentation), retaining the ResNet-50 + spectrogram family. The next few expansions improves the KL further to 
0.516; and continued refinement on the same family, that AIRA abandoned early, ultimately reaches KL 0.467 making \shortname{} the best method on this task.

A few more illustrations of the same pattern are APTOS (a constant-prediction collapse at lr$=10^{-3}$ misattributed by score-only methods to ConvNeXt itself; \shortname{} attributes it to the LR schedule and recovers ConvNeXt to QWK 0.92) and Vesuvius (an 11-channel DeepLabV3 misscored at $F_{0.5}=0.165$ because the validation split was set to zero in the training script; \shortname{} reads the log line and repairs the split rather than discarding the wide-input variant) with trees and node-level diagnoses are in Appendix~\ref{app:failure_attribution}.

\begin{finding}
\shortname{} distributes its expansions across \textbf{diverse hypothesis} families compared to score-based methods that concentrate around a small set of hypotheses.
\end{finding}

Exploring a diverse set of hypotheses is important in open-ended search problems, like ML research. This is because many different approaches can succeed, a new architecture, loss, data representation, or optimizer. The search must explore widely, refining the approach that scored best first is not guaranteed to give global optimal solution. The prior score-only methods concentrate the search to a small set of leading hypotheses and only explore its close variants. In contrast, we find that \shortname{} proposes a higher range of diverse hypotheses, widening the explored set of hypothesis substantially more compared to score-only methods. 

We first validate this by labeling every proposed hypothesis into one of eight axes using a coding agent (GPT-5.5). We find that \shortname{} reaches the widest coverage with \textbf{4.43} axes per run compared to 4.05 for MLEvolve, which is 1.05 more than the number of initial draft nodes. This shows that prior methods like MLEvolve do not explore much beyond the hypotheses proposed in the first step. 

We also measure the entropy of the distribution of a run's proposals over the eight axes. Entropy tells us how evenly those proposals are spread across the axes, so a higher value means the search distributes its budget more uniformly and explores more broadly. We find that \shortname{} produces an entropy of \textbf{1.73}, against 1.35 for MLEvolve and 1.48 for AIRA. This shows that \shortname{} spreads its budget across the axes more evenly than the baselines, which keep returning to a few. 

Next, we perform human validation, we hand-label $111$ expansions across $15$ runs as either a completely new hypothesis, for instance architectural changes, or a minor change within the current hypothesis, like hyperparameter tuning. We find that \textbf{45\%} of \shortname{}'s expansions open a new completely new hypothesis, against only $3\%$ for AIRA and $0\%$ for MLEvolve. This means a massive portion of score-only baselines is exploring minor tweaks within the current hypothesis, limiting the exploration required to solve such tasks.  

Finally, we calculate the within-run pairwise TF-IDF distance between hypotheses. We find that it averages to \textbf{0.72} for \shortname{} and 0.23 for AIRA and 0.14 for MLEvolve, showing that hypotheses explored by \shortname{} are lexically far, whereas for the prior methods, the hypotheses are lexically similar.

To illustrate this diversity qualitatively, we present the Vesuvius ink-detection search trees (Appendix Fig.~\ref{fig:vesuvius_tree}). Comparing \shortname{} and AIRA  we find that \shortname{} holds five distinct axes as siblings at the root (architecture, slice representation, loss, training data, augmentation), identifies that adding the second fragment is a promising strategy (F$_{0.5}=0.479$), and deepens that one branch with augmentation to reach $\mathbf{0.575}$. Whereas, AIRA focuses on tweaking architectures ($3$D U-Net, $2.5$D U-Net with a pretrained encoder, CRNN, sparse $3$D conv, Video-ViT, Swin-UNETR) with a single deep chain.

\begin{table}[t]
  \caption{$\shortname{}^*$: \textbf{test-time training a Qwen3-4B scientist on MLGym tasks.} The Qwen3-4B scientist is test time trained on its own search history; we report mean $\pm$ standard error of the best score over 3 runs, with the best method highlighted. We find that the small 4B scientist match or surpass the o3 scientist on about half the tasks at a fraction of the inference cost.}
  \label{tab:ttt_mlgym}
  \centering
  \setlength{\tabcolsep}{2.2pt}
  \renewcommand{\arraystretch}{1.05}
  \scriptsize
  \resizebox{\textwidth}{!}{%
	  \begin{tabular}{ll ccccccc}
	    \toprule
	    & & \multicolumn{3}{c}{\textit{Prior Works}} & \multicolumn{3}{c}{\textit{Ours}} & \\
	    \cmidrule(lr){3-5} \cmidrule(lr){6-8}
	    \rowcolor{tableheader}
	    Task & Metric & Linear & AIRA & MLEvolve & \shortname{} (o3) & \shortname{} (Qwen 4B) & $\shortname{}^*$ (4B)  & Human Best \\
    \midrule
    Titanic              & acc $\uparrow$     & 0.951$_{\pm.004}$ & 0.944$_{\pm.001}$ & 0.946$_{\pm.002}$ & 0.984$_{\pm.004}$ & 0.949$_{\pm.009}$ & \cellcolor{maxhighlightcolor}\textbf{0.998}$_{\pm.002}$ & 0.830 \\
    CIFAR-10             & acc $\uparrow$     & 0.956$_{\pm.002}$ & 0.964$_{\pm.002}$ & 0.959$_{\pm.003}$ & 0.971$_{\pm.017}$ & 0.957$_{\pm.002}$ & \cellcolor{maxhighlightcolor}\textbf{0.982}$_{\pm.006}$ & 0.994 \\
    Fashion MNIST        & acc $\uparrow$     & 0.946$_{\pm.006}$ & 0.947$_{\pm.000}$ & 0.950$_{\pm.001}$ & \cellcolor{maxhighlightcolor}\textbf{0.958}$_{\pm.001}$ & 0.946$_{\pm.002}$ & 0.948$_{\pm.007}$  & 0.968 \\
    House Price          & $R^2$ $\uparrow$   & 0.940$_{\pm.000}$ & \cellcolor{maxhighlightcolor}\textbf{0.944}$_{\pm.001}$ & 0.943$_{\pm.001}$ & 0.939$_{\pm.003}$ & 0.929$_{\pm.001}$ & 0.940$_{\pm.001}$ & 0.990 \\
    \midrule
    MNLI                 & acc $\uparrow$     & \textbf{84.42}$_{\pm0.05}$ & 83.77$_{\pm0.05}$ & \textbf{84.26}$_{\pm0.08}$ & \cellcolor{maxhighlightcolor}\textbf{84.71}$_{\pm0.49}$ & 83.33$_{\pm.08}$ & 84.01$_{\pm1.81}$ & 92.50 \\
    Lang.\ Modeling      & loss $\downarrow$  & 3.986$_{\pm.169}$ & 4.673$_{\pm.000}$ & 4.015$_{\pm.146}$ & 3.827$_{\pm.088}$ & 4.34$_{\pm.48}$ & \cellcolor{maxhighlightcolor}\textbf{3.518}$_{\pm.276 }$ & 3.500 \\
    \midrule
    Battle of Sexes      & payoff $\uparrow$  & 1.448$_{\pm.001}$ & 1.442$_{\pm.001}$ & 1.446$_{\pm.000}$ & \cellcolor{maxhighlightcolor}\textbf{2.000}$_{\pm.000}$ & 1.441$_{\pm.000}$ & 1.442$_{\pm.000}$ & 1.667 \\
    Prisoner's Dilemma   & payoff $\uparrow$  & 2.453$_{\pm.102}$ & 2.501$_{\pm.129}$ & \textbf{2.635}$_{\pm.012}$ & \cellcolor{maxhighlightcolor}\textbf{2.857}$_{\pm.225}$ & \textbf{2.641}$_{\pm.001}$ & \textbf{2.633}$_{\pm.318}$ & 3.000 \\
    Blotto               & score $\uparrow$   & $-$0.076$_{\pm.327}$ & \textbf{0.247}$_{\pm.003}$ & \textbf{0.250}$_{\pm.001}$ & \textbf{0.249}$_{\pm.001}$ & \textbf{0.251}$_{\pm.002}$ & \cellcolor{maxhighlightcolor}\textbf{0.344}$_{\pm.33}$ & 0.500 \\
    \midrule
    MountainCar          & reward $\uparrow$  & 45.44$_{\pm6.27}$ & 80.82$_{\pm11.75}$ & 84.84$_{\pm6.38}$ & \cellcolor{maxhighlightcolor}\textbf{95.73}$_{\pm.69}$ & 56.65$_{\pm1.69}$ & 91.94$_{\pm2.83}$ & 99.00 \\
    Breakout             & reward $\uparrow$  & 64.03$_{\pm10.75}$ & 57.94$_{\pm3.37}$ & \cellcolor{maxhighlightcolor}\textbf{83.28}$_{\pm3.84}$ & 78.00$_{\pm2.23}$ & 48.82$_{\pm.00}$ & 75.95$_{\pm4.83}$ & 100.00 \\
    Meta Maze            & reward $\uparrow$  & 46.80$_{\pm2.61}$ & 36.42$_{\pm6.47}$ & 45.35$_{\pm3.27}$ & 51.20$_{\pm1.03}$ & 30.35$_{\pm20.67}$ & \cellcolor{maxhighlightcolor}\textbf{53.00}$_{\pm.57}$ & 52.50 \\
    \bottomrule
  \end{tabular}%
  }
\end{table}

\begin{finding}
    \shortname{} spends the exploration budget on \textbf{higher quality hypotheses} compared to score-only methods.
\end{finding}

Exploring many directions helps only when the hypotheses explored are of high quality. Since each hypothesis is an expensive training run, a search algorithm that fills the 8-hour budget with low quality hypothesis cannot reach to a good solution. In \shortname{}, before proposing next hypothesis, the \texttt{scientist} reads  previous nodes' code and training log to identify a specific failure, and proposes a hypothesis that addresses it. Each candidate is therefore well reasoned and high quality, rather than a random mutation of the current best code or an extension of one of the highest-scoring branches. 

To qualitatuvely demonstrate this, we present the trees on MetaMaze
(Appendix Fig.~\ref{fig:metamaze_tree}; baseline reward $15.73$,
partial-observability). The problem requires both preserving 2D maze
geometry at the input and recurrent memory. \textcolor{blue}{\shortname{}
looks at the baseline logs and diagnoses that the baseline sets flattens the input}, vectorising
each 2D maze slice before the network sees it, this destroys the wall and goal geometry. \textcolor{blue}{\shortname{} fixes
the geometry first by using CNN encoder ($23.03$), and richer egocentric
observation ($48.57$) it then combines the input fix
with an LSTM recurrent policy to reach $53.0$, above human
best.} AIRA applies the textbook PO-MDP prior, LSTM,
as its first draft ($48.04$) without inspecting the input pipeline,
then stacks larger nets on the LSTM with improper execution that destabilises the training ($17.5$--$22.3$) eventually leading to the method to discard the LSTM family. MLEvolve never leaves PPO knob mutation (entropy,
\texttt{n\_steps}; mean $45.35$).

 

\subsection{Does test-time training help $\shortname{}^*$ search efficiently?}
\label{subsec:ttt-qualitative}
Table~\ref{tab:ttt_mlgym} demonstrates that performance of \texttt{ARTS} improves using our proposed test-time training procedure. 
On multiple MLGym tasks, test-time training raises the mean normalized score of Qwen3-4B by  $\mathbf{40\%}$ from $0.72$ to $\mathbf{1.01}$. 
\textcolor{blue}{A Qwen3-4B scientist updated with test-time training surpass performance of o3-scientist} on half of the tasks and is close on other tasks.
We observe prominent performance gains on tasks of language modeling, where the loss improves from $4.34$ to $\mathbf{3.52}$ and attains human performance, and on MetaMaze where our approache exceeds human performance. 

Through our qualitative analysis, we find that TTT helps when search finds a good direction but the scientist stops using it after long trajectories. For instance, in MetaMaze, which has a partially observable state and requires memory based solutions, prior algorithms propose LSTM/GRU based policy architectures.
Since LSTM/GRU can be unstable with initial implementations, prior appoaches discards them during the search process.
In contrast, \shortname{} with an o3 scientist identifies the need for memory but switches to high-scoring MLP/PPO.
However, Qwen3-4B scientist~(TTT) reconsiders the recurrent-memory direction and derives a well optimized LSTM-based solution. \textcolor{blue}{This suggests that TTT instills useful search experience in the scientist's weights and reduces the dependence on increasingly long context for extended search.} Figure~\ref{fig:metamaze_main} visualises the three trajectories side by side.

\subsection{Ablations.}
\label{subsec:ablations}

\begin{figure}[t]
    \centering
    \includegraphics[width=\linewidth]{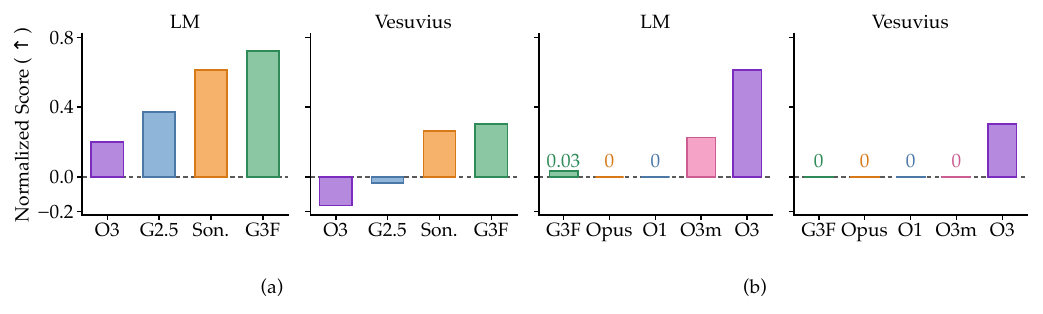}
    \vspace{-2.4em}
    \caption{Swap ablations. (a) Executor swap with the scientist fixed to o3. (b) Scientist swap with the executor fixed to Gemini-3-Flash. The dashed line marks normalized score 0. Small labels show near-zero normalized scores.}
    \label{fig:swap_ablations}
\end{figure}

We study the impact of crucial components of \shortname{}: the executor model, the scientist model, token and executor-call usage, and the search components audit, verbalized sampling, memory, and initial breadth. Fig.~\ref{fig:swap_ablations} shows that the executor matters even when the scientist is fixed: normalized score rises from $0.20$ to $0.72$ on Language Modeling and from $-0.16$ to $0.30$ on Vesuvius. With the executor fixed, o3 is the only scientist that reliably moves beyond baseline on these two hard tasks. \shortname{} also uses fewer estimated tokens than the baselines, $0.72$M versus $0.83$M for AIRA, $1.53$M for Linear, and $2.30$M for MLEvolve. Removing audit or verbalized sampling gives the largest component drops, from $0.809$ to $0.470/0.506$ on HMS and from $0.887$ to $0.573/0.614$ on Kuzushiji (Appendix Fig.~\ref{fig:token_component_ablations}). Full numbers are in Appendix~\ref{sec:extended_ablations}.

For \shortname{} with TTT, we ablate the reward function, the GRPO episode structure, and whether training collapses diversity. The final percentile reward is best on all three reward-ablation tasks, with mean normalized score \textbf{1.03} versus 0.29 for the strongest alternative. Single-step GRPO improves mean normalized score from 0.89 to \textbf{1.48} over tree-per-episode rollouts. Training also preserves diversity: runs cover 6--10 strategy categories per task, and entropy increases from early to late rollouts on 3 of 4 measured tasks. Appendix Figs.~\ref{fig:reward_function_ablation} and~\ref{fig:ttt_diversity_ablation} give the TTT ablations.

\section{Discussion}

\shortname{} replaces heuristic parent selection with reasoning over code, logs, scores, and memory. The qualitative analysis suggests that this helps the agent diagnose why experiments fail, preserve promising directions after weak executions, and open new axes when the current tree is narrow. Across 22 tasks, \shortname{} gives the best automated-search result on 16 tasks, improves normalized score by $15.3\%$ over the strongest baseline, and has the best IQM, $0.93$ versus $0.87$ for MLEvolve. It also explores more broadly, with $4.43$ axes per run and entropy $1.73$. Test-time training raises Qwen3-4B from $0.72$ to $\mathbf{1.01}$ mean normalized score on MLGym.

\para{Limitations and Ethical Concerns}
\label{sec:limitations}
\shortname{} depends on the quality of its scientist and executor, though the modular design lets stronger reasoning and coding models be swapped in independently. Stronger research agents may also concentrate access among groups with more compute or be misused in unsafe domains; deployment should retain task constraints, logging, and human oversight.

\begin{ack}
AJ is supported by Fonds de Recherche du Quebec (FRQ) (DOI assigned: https://doi.org/10.69777/350253),
Calcul Quebec, and Canada Excellence Research Chairs (CERC) program.
The research was enabled in part by computational resources provided by the Digital Research Alliance of Canada
(https://alliancecan.ca) and Mila (https://mila.quebec).

We thank Roberta Raileanu for her early feedback on the draft, and Himanshu Gaurav Singh (UC Berkeley) for useful discussions.
\end{ack}
\newpage


\bibliographystyle{plainnat}
\bibliography{references}

\newpage
\appendix

\section*{Appendix Contents}
\addcontentsline{toc}{section}{Appendix Contents}
\begingroup
\setlength{\parskip}{0.35em}
\noindent\hyperref[app:hms_search_tree]{\textbf{\ref*{app:hms_search_tree}}\quad HMS Search Trajectory: \shortname{} vs.\ AIRA}\dotfill \pageref{app:hms_search_tree}\par
\noindent\hyperref[app:vesuvius_tree]{\textbf{\ref*{app:vesuvius_tree}}\quad Vesuvius Search Trajectory: \shortname{} vs.\ AIRA}\dotfill \pageref{app:vesuvius_tree}\par
\noindent\hyperref[app:metamaze_search_tree]{\textbf{\ref*{app:metamaze_search_tree}}\quad MetaMaze Search Trajectory: \shortname{} vs.\ AIRA}\dotfill \pageref{app:metamaze_search_tree}\par
\noindent\hyperref[app:mountaincar_tree]{\textbf{\ref*{app:mountaincar_tree}}\quad MountainCar Search Trajectory: \shortname{} vs.\ AIRA}\dotfill \pageref{app:mountaincar_tree}\par
\noindent\hyperref[llm]{\textbf{\ref*{llm}}\quad LLM Use}\dotfill \pageref{llm}\par
\noindent\hyperref[app:qualitative_search_efficiency]{\textbf{\ref*{app:qualitative_search_efficiency}}\quad Qualitative Analysis of Search Efficiency}\dotfill \pageref{app:qualitative_search_efficiency}\par
\noindent\hyperref[app:failure_attribution]{\textbf{\ref*{app:failure_attribution}}\quad Failure Attribution: Detailed Examples}\dotfill \pageref{app:failure_attribution}\par
\noindent\hyperref[app:extended_related_work]{\textbf{\ref*{app:extended_related_work}}\quad Extended Related Work}\dotfill \pageref{app:extended_related_work}\par
\noindent\hyperref[app:experimental_details]{\textbf{\ref*{app:experimental_details}}\quad Additional Experimental Details}\dotfill \pageref{app:experimental_details}\par
\noindent\hyperref[app:references_to_human_best]{\textbf{\ref*{app:references_to_human_best}}\quad Sources for Human-Best Scores}\dotfill \pageref{app:references_to_human_best}\par
\noindent\hyperref[app:scientist_prompts]{\textbf{\ref*{app:scientist_prompts}}\quad Scientist Prompts}\dotfill \pageref{app:scientist_prompts}\par
\noindent\hyperref[sec:extended_ablations]{\textbf{\ref*{sec:extended_ablations}}\quad Extended Ablations}\dotfill \pageref{sec:extended_ablations}\par
\endgroup

\vspace{1.5em}

The four search-tree figures (\S\ref{app:hms_search_tree}--\S\ref{app:mountaincar_tree}) are referenced from Findings~1--3 in the main text and visualise the full per-node trajectories described there. The remaining appendices (\S\ref{app:qualitative_search_efficiency}--\S\ref{sec:extended_ablations}) collect supporting analyses, prompts, experimental details, and ablation tables.

\newpage
\section{HMS Search Trajectory: \shortname{} versus AIRA}
\label{app:hms_search_tree}

\begin{figure}[H]
    \centering
    \resizebox{\linewidth}{!}{%
    \begin{tikzpicture}[
        font=\scriptsize,
        nodebar/.style={
            draw=#1!70!black,
            fill=#1!14,
            rounded corners=2pt,
            line width=0.65pt,
            align=left,
            anchor=north west,
            text width=6.4cm,
            inner xsep=4pt,
            inner ysep=3pt
        },
        bestbar/.style={
            draw=#1!90!black,
            fill=#1!22,
            rounded corners=2pt,
            line width=1.5pt,
            align=left,
            anchor=north west,
            text width=6.4cm,
            inner xsep=4pt,
            inner ysep=3pt
        },
        sidenote/.style={
            align=left,
            anchor=north west,
            text width=6.0cm,
            inner xsep=2pt,
            inner ysep=2pt,
            font=\fontsize{6.5pt}{7.8pt}\selectfont\itshape,
            text=gray!25!black
        },
        rowbox/.style={fit=#1, draw=none, inner sep=0pt},
        connector/.style={draw=gray!72, line width=0.7pt, rounded corners=1.5pt},
        sectionhdr/.style={anchor=north west, font=\scriptsize\bfseries}
    ]
    \node[sectionhdr] (h1) at (0,0) {\shortname{} \;-\; 18 nodes, depth 6, best KL $=$ \textbf{0.467}};
    \node[nodebar=deepblue] (n_r) at ([yshift=-2mm]h1.south west)
        {\texttt{[root]} uniform-prediction baseline, KL $1.462$};
    \node[sidenote] (sn_r) at ([xshift=3mm]n_r.north east)
        {step 0 \texttt{INSPECT NONE}: baseline is the trivial uniform predictor; data are EEG spectrograms with patient-level grouping $\rightarrow$ draft $5$ families in parallel covering architecture, data, and augmentation axes};
    \node[rowbox={(n_r)(sn_r)}] (row_r) {};

    \node[nodebar=deeppurple, xshift=3mm] (n_0) at ([yshift=-1.5mm]row_r.south west)
        {\texttt{[root\_0, arch]} ResNet-18, 1-channel log1p spectrogram, KL $0.623$};
    \node[sidenote] (sn_0) at ([xshift=3mm]n_0.north east)
        {initial breadth $\#1$: shallow single-channel backbone as the cheap reference family};
    \node[rowbox={(n_0)(sn_0)}] (row_0) {};
    \draw[connector] ([xshift=2mm]n_r.south west) |- (n_0.west);

    \node[nodebar=deeppurple, xshift=3mm] (n_1) at ([yshift=-1.5mm]row_0.south west)
        {\texttt{[root\_1, data]} ResNet-18 $+$ SpecAugment, KL $0.562$};
    \node[sidenote] (sn_1) at ([xshift=3mm]n_1.north east)
        {initial breadth $\#2$: augmentation axis kept alive in parallel};
    \node[rowbox={(n_1)(sn_1)}] (row_1) {};
    \draw[connector] ([xshift=2mm]n_r.south west) |- (n_1.west);

    \node[bestbar=deepteal, xshift=3mm] (n_2) at ([yshift=-1.5mm]row_1.south west)
        {\texttt{[root\_2, arch]} \textbf{pretrained ResNet-50, 3-channel spectrogram} (C0 log1p, C1 temporal-diff), KL $\mathbf{0.557}$};
    \node[sidenote] (sn_2) at ([xshift=3mm]n_2.north east)
        {step 1 \texttt{INSPECT root\_0, root\_1}: depth-$1$ winner; \textbf{larger pretrained backbone + multi-channel spectrogram} is the strongest direction $\rightarrow$ deepen this family};
    \node[rowbox={(n_2)(sn_2)}] (row_2) {};
    \draw[connector] ([xshift=2mm]n_r.south west) |- (n_2.west);

    \node[nodebar=BrickRed, xshift=3mm] (n_20) at ([yshift=-1.5mm]row_2.south west)
        {\texttt{[root\_2\_0, arch]} swap to EfficientNet-B0 ($3$-ch), KL $\mathbf{0.702}$ \;\textbf{(regression $+0.145$)}};
    \node[sidenote] (sn_20) at ([xshift=3mm]n_20.north east)
        {step 2 \texttt{INSPECT root\_2\_0}: backbone swap regressed despite identical input pipeline. Val curve still falling at last epoch $\rightarrow$ EfficientNet \textbf{under-trained} at default schedule, not a failure of the spectrogram family};
    \node[rowbox={(n_20)(sn_20)}] (row_20) {};
    \draw[connector] ([xshift=2mm]n_2.south west) |- (n_20.west);

    \node[nodebar=BrickRed, xshift=3mm] (n_21) at ([yshift=-1.5mm]row_20.south west)
        {\texttt{[root\_2\_1, comb]} ResNet-50 $+$ SpecAugment, KL $\mathbf{0.565}$ \;\textbf{(slight regression $+0.008$)}};
    \node[sidenote] (sn_21) at ([xshift=3mm]n_21.north east)
        {step 3 \texttt{INSPECT root\_2\_1}: \textbf{two consecutive regressions on the same family}. Logs show SpecAugment injects augmentation noise that slows optimisation; classify as \textbf{implementation-wrong (under-trained under heavy augmentation)} $\rightarrow$ \textbf{retain the family}, try a different augmentation};
    \node[rowbox={(n_21)(sn_21)}] (row_21) {};
    \draw[connector] ([xshift=2mm]n_2.south west) |- (n_21.west);

    \node[nodebar=deeppurple, xshift=3mm] (n_22) at ([yshift=-1.5mm]row_21.south west)
        {\texttt{[root\_2\_2, data]} ResNet-50 $+$ Mixup (Beta$(0.4,0.4)$), KL $0.527$};
    \node[sidenote] (sn_22) at ([xshift=3mm]n_22.north east)
        {step 4 \texttt{INSPECT root\_2\_2}: \textbf{first improvement over the parent} from the same family; Mixup counters over-confidence on noisy EEG labels};
    \node[rowbox={(n_22)(sn_22)}] (row_22) {};
    \draw[connector] ([xshift=2mm]n_2.south west) |- (n_22.west);

    \node[nodebar=deeppurple, xshift=6mm] (n_222) at ([yshift=-1.5mm]row_22.south west)
        {\texttt{[root\_2\_2\_2, comb]} Mixup $+$ SpecAugment stacked, KL $0.519$};
    \node[sidenote] (sn_222) at ([xshift=3mm]n_222.north east)
        {step 5 \texttt{INSPECT root\_2\_2\_\{0,1,2\}}: Mixup $+$ SpecAugment applied on a strictly better base than at \texttt{root\_2\_1}; ResNet-101 swap (\texttt{..\_2\_2\_0}) regressed to $0.542$ -- backbone capacity not the bottleneck};
    \node[rowbox={(n_222)(sn_222)}] (row_222) {};
    \draw[connector] ([xshift=2mm]n_22.south west) |- (n_222.west);

    \node[nodebar=deeppurple, xshift=6mm] (n_2220) at ([yshift=-1.5mm]row_222.south west)
        {\texttt{[root\_2\_2\_2\_0, data]} $+$ per-channel normalisation ($\mu,\sigma$ over $10{,}024$ training spectrograms), KL $0.516$};
    \node[sidenote] (sn_2220) at ([xshift=3mm]n_2220.north east)
        {step 6 \texttt{READ train\_and\_predict.py}: spectrograms used unnormalised across channels; compute dataset-level statistics ($\mu_{C_0}\!=\!1.10, \mu_{C_1}\!=\!0.74, \ldots$) and standardise};
    \node[rowbox={(n_2220)(sn_2220)}] (row_2220) {};
    \draw[connector] ([xshift=2mm]n_222.south west) |- (n_2220.west);

    \node[nodebar=deeppurple, xshift=6mm] (n_22204) at ([yshift=-1.5mm]row_2220.south west)
        {\texttt{[root\_2\_2\_2\_0\_4, comb]} $+$ custom 6-class FC head, 5-fold patient-level GroupKFold, KL $0.485$};
    \node[sidenote] (sn_22204) at ([xshift=3mm]n_22204.north east)
        {steps 7-8 \texttt{INSPECT siblings of} \texttt{..\_0}: per-channel norm fixed, but $\#$folds and head capacity still on default $\rightarrow$ replace default head with a dedicated 6-class \texttt{Linear(2048,6)} layer and bump to $5$-fold GroupKFold by \texttt{patient\_id}};
    \node[rowbox={(n_22204)(sn_22204)}] (row_22204) {};
    \draw[connector] ([xshift=2mm]n_2220.south west) |- (n_22204.west);

    \node[bestbar=deepteal, xshift=9mm] (n_222041) at ([yshift=-1.5mm]row_22204.south west)
        {\texttt{[root\_2\_2\_2\_0\_4\_1, hp]} $+$ tuned LR / $8$-epoch schedule, \textbf{KL $\mathbf{0.467}$} \;{\large$\bigstar$}};
    \node[sidenote] (sn_222041) at ([xshift=3mm]n_222041.north east)
        {step 9 \texttt{INSPECT root\_2\_2\_2\_0\_4}: epoch-level curves show $8$-epoch schedule is the right horizon; finer LR tuning closes the remaining gap $\rightarrow$ \textbf{best KL on the task}};
    \node[rowbox={(n_222041)(sn_222041)}] (row_222041) {};
    \draw[connector] ([xshift=2mm]n_22204.south west) |- (n_222041.west);

    \node[sectionhdr] (h2) at ([yshift=-5mm, xshift=-21mm]row_222041.south west)
        {AIRA \;-\; 29 nodes, depth 5, best KL $=$ \textbf{0.513} (draft $1$; never refined)};
    \node[nodebar=deepblue] (a_r) at ([yshift=-2mm]h2.south west)
        {\texttt{[root]} uniform-prediction baseline, KL $1.462$};
    \node[sidenote] (asn_r) at ([xshift=3mm]a_r.north east)
        {UCB selection only -- no \texttt{scientist}; next node is whichever leaf maximises UCB};
    \node[rowbox={(a_r)(asn_r)}] (arow_r) {};

    \node[bestbar=deepteal, xshift=3mm] (a_0) at ([yshift=-1.5mm]arow_r.south west)
        {\texttt{[root\_0]} draft $1$: ResNet-style spectrogram pipeline, KL $\mathbf{0.513}$ {\large$\bigstar$} \;(\textbf{abandoned})};
    \node[sidenote] (sn_a0) at ([xshift=3mm]a_0.north east)
        {AIRA's $\textsc{Inspect}$ does not classify the failure mode of subsequent regressions; UCB picks the highest-mean leaf so the $0.513$ family is never revisited after the next two siblings score above it};
    \node[rowbox={(a_0)(sn_a0)}] (arow_0) {};
    \draw[connector] ([xshift=2mm]a_r.south west) |- (a_0.west);

    \node[nodebar=BrickRed, xshift=3mm] (a_1) at ([yshift=-1.5mm]arow_0.south west)
        {\texttt{[root\_1]} draft $2$: hand-crafted features via \texttt{scipy.signal} (skew, kurtosis, band-power), KL $0.712$};
    \node[sidenote] (sn_a1) at ([xshift=3mm]a_1.north east)
        {first pivot: drops the spectrogram-CNN family entirely};
    \node[rowbox={(a_1)(sn_a1)}] (arow_1) {};
    \draw[connector] ([xshift=2mm]a_r.south west) |- (a_1.west);

    \node[nodebar=BrickRed, xshift=3mm] (a_2) at ([yshift=-1.5mm]arow_1.south west)
        {\texttt{[root\_2]} draft $3$: derivative-stacked spectrogram $+$ EfficientNet, KL $0.662$};
    \node[sidenote] (sn_a2) at ([xshift=3mm]a_2.north east)
        {second pivot: yet another family; UCB then deepens \texttt{root\_2} because it has fewer children};
    \node[rowbox={(a_2)(sn_a2)}] (arow_2) {};
    \draw[connector] ([xshift=2mm]a_r.south west) |- (a_2.west);

    \node[nodebar=BrickRed, xshift=6mm] (a_20) at ([yshift=-1.5mm]arow_2.south west)
        {\texttt{[root\_2\_0]} EfficientNet variant with 3-channel mismatch, \textbf{KL $\mathbf{4.080}$} \;(catastrophic)};
    \node[sidenote] (sn_a20) at ([xshift=3mm]a_20.north east)
        {bug in channel handling sends KL $\sim\!3\!\times$ above the trivial baseline -- but UCB still deepens this branch};
    \node[rowbox={(a_20)(sn_a20)}] (arow_20) {};
    \draw[connector] ([xshift=2mm]a_2.south west) |- (a_20.west);

    \node[nodebar=BrickRed, xshift=9mm] (a_200) at ([yshift=-1.5mm]arow_20.south west)
        {\texttt{[root\_2\_0\_0]} compounded variant, \textbf{KL $\mathbf{9.430}$} \;(diverges further)};
    \node[sidenote] (sn_a200) at ([xshift=3mm]a_200.north east)
        {a working family ($0.513$) abandoned after one node is a working family the search never refines};
    \node[rowbox={(a_200)(sn_a200)}] (arow_200) {};
    \draw[connector] ([xshift=2mm]a_20.south west) |- (a_200.west);
    \end{tikzpicture}%
    }
    \caption{HMS Brain Activity search trees. \shortname{} (top)
    drafts a pretrained ResNet-50 over 3-channel log-mel spectrograms
    at depth $1$ and reaches KL $\mathbf{0.557}$. The next two
    children regress (EfficientNet-B0, $0.702$; SpecAugment, $0.565$).
    \shortname{} re-reads the regressed nodes' training logs, classifies
    them as \emph{implementation-wrong} (under-trained under heavy
    augmentation), retains the ResNet-50 + spectrogram family, and the
    next several expansions compound Mixup, SpecAugment, per-channel
    normalisation, a 5-fold patient-level GroupKFold, and an LR /
    schedule sweep to reach \textbf{KL $\mathbf{0.467}$} -- the best
    on the task. AIRA (bottom) reaches a comparable family at its
    first draft ($0.513$) but its next two drafts pivot to hand-crafted
    features and a derivative-stacked EfficientNet; UCB then deepens
    the latter pivot whose children diverge catastrophically to KL
    $\mathbf{4.08}$ and $\mathbf{9.43}$, and AIRA never returns to the
    original $0.513$ family.}
    \label{fig:hms_tree}
\end{figure}
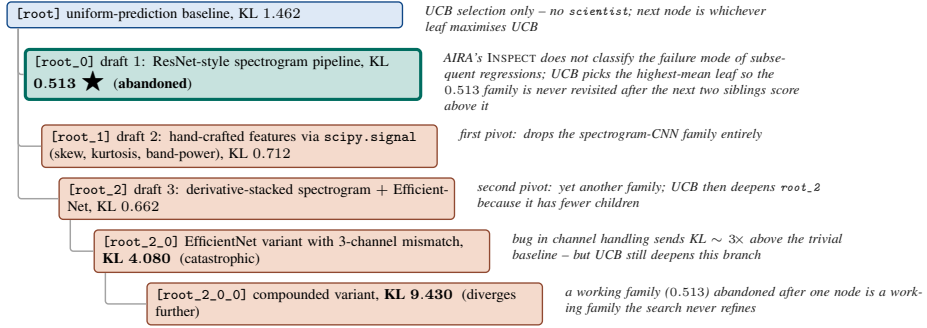

We contrast \shortname{} and AIRA on HMS Harmful Brain Activity
Classification (MLE-bench; KL divergence, \emph{lower} is better;
uniform-prediction baseline KL $1.462$). Tree nodes are coloured
\textcolor{deepblue!70!black}{\textbf{blue}} (baseline),
\textcolor{deeppurple!70!black}{\textbf{purple}} (progress),
\textcolor{deepteal!70!black}{\textbf{teal}} (new best, $\bigstar$),
and \textcolor{BrickRed!80!black}{\textbf{red}}
(regression / abandoned family). Italic margin notes give the
scientist's tool call (\texttt{INSPECT}/\texttt{READ}) at that step
together with the reasoning that drove the next experiment.
\shortname{} reaches best KL $\mathbf{0.467}$; AIRA pivots after its
first draft (KL $0.513$) and the children of its pivot diverge to KL
$4.08$ and $9.43$, never returning to the working family.

\section{Vesuvius Search Trajectory: \shortname{} versus AIRA}
\label{app:vesuvius_tree}

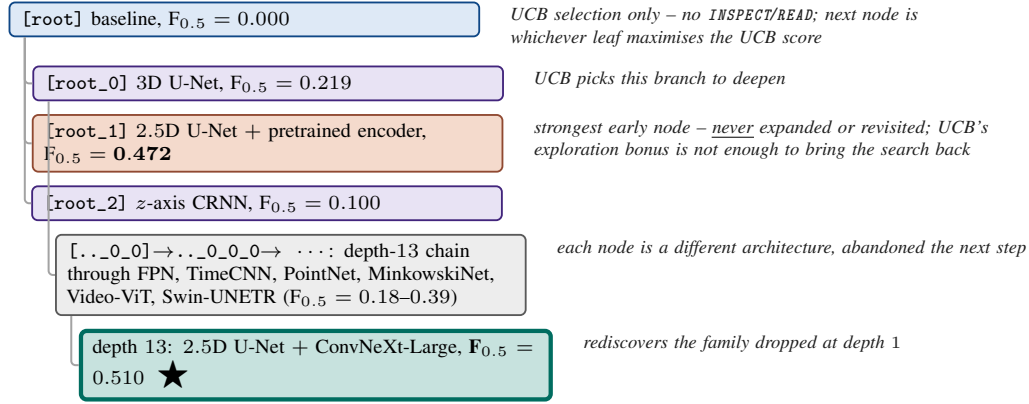
\begin{figure}
    \centering
    \resizebox{\linewidth}{!}{%
    \begin{tikzpicture}[
        font=\scriptsize,
        nodebar/.style={
            draw=#1!70!black,
            fill=#1!14,
            rounded corners=2pt,
            line width=0.65pt,
            align=left,
            anchor=north west,
            text width=5.5cm,
            inner xsep=4pt,
            inner ysep=3pt
        },
        bestbar/.style={
            draw=#1!90!black,
            fill=#1!22,
            rounded corners=2pt,
            line width=1.5pt,
            align=left,
            anchor=north west,
            text width=5.5cm,
            inner xsep=4pt,
            inner ysep=3pt
        },
        sidenote/.style={
            align=left,
            anchor=north west,
            text width=5.8cm,
            inner xsep=2pt,
            inner ysep=2pt,
            font=\fontsize{6.5pt}{7.8pt}\selectfont\itshape,
            text=gray!25!black
        },
        rowbox/.style={fit=#1, draw=none, inner sep=0pt},
        connector/.style={draw=gray!72, line width=0.7pt, rounded corners=1.5pt},
        sectionhdr/.style={anchor=north west, font=\scriptsize\bfseries}
    ]
    \node[sectionhdr] (h1) at (0,0) {\shortname{} \;-\; 13 nodes, depth 3, best F$_{0.5}=$ \textbf{0.575}};
    \node[nodebar=deepblue] (n_r) at ([yshift=-2mm]h1.south west) {\texttt{[root]} baseline, F$_{0.5}=0.000$};
    \node[sidenote] (sn_r) at ([xshift=3mm]n_r.north east) {step 0 \texttt{READ baseline.py}: baseline keeps $10$ of $65$ slices and \emph{averages} them, \textbf{discarding the $3$D structure entirely} $\rightarrow$ no downstream tweak can recover the signal once averaged $\rightarrow$ \textbf{first move is a CNN} that uses the slices directly};
    \node[rowbox={(n_r)(sn_r)}] (row_r) {};
    \node[nodebar=deeppurple, xshift=3mm] (n_0) at ([yshift=-1.5mm]row_r.south west) {\texttt{[root\_0, arch]} DeepLabV3-ResNet50, $3$-slice, F$_{0.5}=0.151$};
    \node[sidenote] (sn_0) at ([xshift=3mm]n_0.north east) {step 1 \texttt{INSPECT root\_0}: \textbf{ink lives a few microns above/below the surface}; $3$-slice context is too narrow ($62/65$ slices still discarded) $\rightarrow$ \textbf{lift the input to $11$ slices} ($z\!-\!5 \ldots z\!+\!5$)};
    \node[rowbox={(n_0)(sn_0)}] (row_0) {};
    \draw[connector] ([xshift=2mm]n_r.south west) |- (n_0.west);
    \node[nodebar=deeppurple] (n_1) at ([yshift=-1.5mm]row_0.south west) {\texttt{[root\_1, data]} $11$-channel slice stack, F$_{0.5}=0.165$};
    \node[sidenote] (sn_1) at ([xshift=3mm]n_1.north east) {step 2 \texttt{INSPECT root\_1}: \textbf{BCE on $<\!1\%$ ink pixels} encourages predict-background-everywhere $\rightarrow$ \textbf{switch to Focal $+$ BCE}};
    \node[rowbox={(n_1)(sn_1)}] (row_1) {};
    \draw[connector] ([xshift=2mm]n_r.south west) |- (n_1.west);
    \node[nodebar=deeppurple] (n_2) at ([yshift=-1.5mm]row_1.south west) {\texttt{[root\_2, loss]} Focal $+$ BCE, F$_{0.5}=0.034$ (diverged)};
    \node[sidenote] (sn_2) at ([xshift=3mm]n_2.north east) {step 3 \texttt{INSPECT root\_1}: pipeline trains on \emph{fragment 1 only} with $4{,}000$ random patches; \textbf{fragment $2$ is fully labelled and unused} $\rightarrow$ \textbf{doubling the labelled data} is the cheapest gain};
    \node[rowbox={(n_2)(sn_2)}] (row_2) {};
    \draw[connector] ([xshift=2mm]n_r.south west) |- (n_2.west);
    \node[bestbar=deepteal] (n_3) at ([yshift=-1.5mm]row_2.south west) {\texttt{[root\_3, data]} train on fragments $1\,\&\,2$, F$_{0.5}=\mathbf{0.479}$ best of breadth};
    \node[sidenote] (sn_3) at ([xshift=3mm]n_3.north east) {step 4 \texttt{INSPECT root\_3}: \texttt{root\_3} is good but \textbf{augmentation completely untested} ($0$ attempts) $\rightarrow$ \textbf{probe aug as a sibling first}; if aug breaks the model we still have \texttt{root\_3} to deepen};
    \node[rowbox={(n_3)(sn_3)}] (row_3) {};
    \draw[connector] ([xshift=2mm]n_r.south west) |- (n_3.west);
    \node[nodebar=deeppurple] (n_4) at ([yshift=-1.5mm]row_3.south west) {\texttt{[root\_4, data]} strong aug ($\pm45^\circ$, photometric), F$_{0.5}=0.105$ (collapsed)};
    \node[sidenote] (sn_4) at ([xshift=3mm]n_4.north east) {step 6 \texttt{INSPECT root\_3}: \texttt{root\_4}'s \textbf{aggressive aug killed the signal} $\rightarrow$ apply only the \textbf{conservative subset (flips $+$ $90^\circ$ rotations)} on the $0.479$ winner};
    \node[rowbox={(n_4)(sn_4)}] (row_4) {};
    \draw[connector] ([xshift=2mm]n_r.south west) |- (n_4.west);
    \node[bestbar=deepteal, xshift=3mm] (n_31) at ([yshift=-1.5mm]row_4.south west) {\texttt{[root\_3\_1, data]} flips $+$ $90^\circ$ rotations, \textbf{F$_{0.5}=0.575$} \;{\large$\bigstar$}};
    \node[sidenote] (sn_31) at ([xshift=3mm]n_31.north east) {steps 7-12 \texttt{INSPECT root\_3\_1} (occasionally also \texttt{root\_3}): refine around the winner -- inference-side levers (TTA, threshold, sampler) and a final loss-reweighting sweep};
    \node[rowbox={(n_31)(sn_31)}] (row_31) {};
    \draw[connector] ([xshift=2mm]n_3.south west) |- (n_31.west);
    \node[nodebar=gray, xshift=3mm] (n_ref) at ([yshift=-1.5mm]row_31.south west) {\texttt{[root\_3\_1\_\{0..4\}]} TTA, thresh, sampler, \texttt{pos\_weight}, $3$-ch: F$_{0.5}=0.05$--$0.50$};
    \node[sidenote] (sn_ref) at ([xshift=3mm]n_ref.north east) {none beats $0.575$ within budget; remaining gains likely require a stronger backbone};
    \node[rowbox={(n_ref)(sn_ref)}] (row_ref) {};
    \draw[connector] ([xshift=2mm]n_31.south west) |- (n_ref.west);

    \node[sectionhdr] (h2) at ([yshift=-5mm, xshift=-8mm]row_ref.south west) {AIRA \;-\; depth 13, best F$_{0.5}=$ \textbf{0.510}, no scientist tool calls};
    \node[nodebar=deepblue] (a_r) at ([yshift=-2mm]h2.south west) {\texttt{[root]} baseline, F$_{0.5}=0.000$};
    \node[sidenote] (asn_r) at ([xshift=3mm]a_r.north east) {UCB selection only -- no \texttt{INSPECT}/\texttt{READ}; next node is whichever leaf maximises the UCB score};
    \node[rowbox={(a_r)(asn_r)}] (arow_r) {};
    \node[nodebar=deeppurple, xshift=3mm] (a_0) at ([yshift=-1.5mm]arow_r.south west) {\texttt{[root\_0]} $3$D U-Net, F$_{0.5}=0.219$};
    \node[sidenote] (asn_0) at ([xshift=3mm]a_0.north east) {UCB picks this branch to deepen};
    \node[rowbox={(a_0)(asn_0)}] (arow_0) {};
    \draw[connector] ([xshift=2mm]a_r.south west) |- (a_0.west);
    \node[nodebar=BrickRed] (a_1) at ([yshift=-1.5mm]arow_0.south west) {\texttt{[root\_1]} $2.5$D U-Net $+$ pretrained encoder, F$_{0.5}=\mathbf{0.472}$};
    \node[sidenote] (asn_1) at ([xshift=3mm]a_1.north east) {strongest early node -- \emph{never} expanded or revisited; UCB's exploration bonus is not enough to bring the search back};
    \node[rowbox={(a_1)(asn_1)}] (arow_1) {};
    \draw[connector] ([xshift=2mm]a_r.south west) |- (a_1.west);
    \node[nodebar=deeppurple] (a_2) at ([yshift=-1.5mm]arow_1.south west) {\texttt{[root\_2]} $z$-axis CRNN, F$_{0.5}=0.100$};
    \node[sidenote] (asn_2) at ([xshift=3mm]a_2.north east) {};
    \node[rowbox={(a_2)(asn_2)}] (arow_2) {};
    \draw[connector] ([xshift=2mm]a_r.south west) |- (a_2.west);
    \node[nodebar=gray, xshift=3mm] (a_ch) at ([yshift=-1.5mm]arow_2.south west) {\texttt{[..\_0\_0]}$\rightarrow$\texttt{..\_0\_0\_0}$\rightarrow\cdots$: depth-$13$ chain through FPN, TimeCNN, PointNet, MinkowskiNet, Video-ViT, Swin-UNETR (F$_{0.5}=0.18$--$0.39$)};
    \node[sidenote] (asn_ch) at ([xshift=3mm]a_ch.north east) {each node is a different architecture, abandoned the next step};
    \node[rowbox={(a_ch)(asn_ch)}] (arow_ch) {};
    \draw[connector] ([xshift=2mm]a_0.south west) |- (a_ch.west);
    \node[bestbar=deepteal, xshift=3mm] (a_w) at ([yshift=-1.5mm]arow_ch.south west) {depth $13$: $2.5$D U-Net $+$ ConvNeXt-Large, \textbf{F$_{0.5}=0.510$} \;{\large$\bigstar$}};
    \node[sidenote] (asn_w) at ([xshift=3mm]a_w.north east) {rediscovers the family dropped at depth $1$};
    \node[rowbox={(a_w)(asn_w)}] (arow_w) {};
    \draw[connector] ([xshift=2mm]a_ch.south west) |- (a_w.west);
    \end{tikzpicture}%
    }
    \caption{Vesuvius Ink Detection search trees. \shortname{} (top) keeps
    five distinct axes as siblings of \texttt{root}, deepens the dominant
    lever (\texttt{root\_3}, both fragments, $0.479$) with augmentation to
    $\mathbf{0.575}$. AIRA (bottom) commits to the $3$D-U-Net branch via UCB,
    abandons a stronger $2.5$D-pretrained sibling at depth $1$, and only
    rediscovers that family $12$ nodes deeper at $0.510$.}
    \label{fig:vesuvius_tree}
\end{figure}

We contrast \shortname{} and AIRA on Vesuvius Challenge Ink Detection
(MLE-bench; F$_{0.5}$, higher is better). Tree nodes are colored
\textcolor{deepblue!70!black}{\textbf{blue}} (baseline),
\textcolor{deeppurple!70!black}{\textbf{purple}} (progress),
\textcolor{deepteal!70!black}{\textbf{teal}} (new best, $\bigstar$),
\textcolor{BrickRed!80!black}{\textbf{red}} (regression / abandoned strong
node). Depth is shown by indentation, \texttt{par=X} labels, and gray
parent$\rightarrow$child L-branches. Italic margin notes give the
scientist's tool call and the reasoning that drove the next experiment.

\section{MetaMaze Search Trajectory: \shortname{} versus AIRA}
\label{app:metamaze_search_tree}
\begin{figure}[H]
    \centering
    \resizebox{\linewidth}{!}{%
    \begin{tikzpicture}[
        font=\scriptsize,
        nodebar/.style={
            draw=#1!70!black,
            fill=#1!14,
            rounded corners=2pt,
            line width=0.65pt,
            align=left,
            anchor=north west,
            text width=5.5cm,
            inner xsep=4pt,
            inner ysep=3pt
        },
        bestbar/.style={
            draw=#1!90!black,
            fill=#1!22,
            rounded corners=2pt,
            line width=1.5pt,
            align=left,
            anchor=north west,
            text width=5.5cm,
            inner xsep=4pt,
            inner ysep=3pt
        },
        sidenote/.style={
            align=left,
            anchor=north west,
            text width=5.8cm,
            inner xsep=2pt,
            inner ysep=2pt,
            font=\fontsize{6.5pt}{7.8pt}\selectfont\itshape,
            text=gray!25!black
        },
        rowbox/.style={fit=#1, draw=none, inner sep=0pt},
        connector/.style={draw=gray!72, line width=0.7pt, rounded corners=1.5pt},
        sectionhdr/.style={anchor=north west, font=\scriptsize\bfseries}
    ]
    \node[sectionhdr] (h1) at (0,0) {\shortname{} \;-\; 7 nodes, depth 4, best reward $=$ \textbf{48.57}. Scientist proposes LSTM/GRU but verbalized sampler never selects it.};
    \node[nodebar=deepblue] (n_r) at ([yshift=-2mm]h1.south west) {\texttt{[root]} \texttt{flatten\_3d=True} $+$ MLP PPO baseline, reward $15.73$};
    \node[sidenote] (sn_r) at ([xshift=3mm]n_r.north east) {step 0 \texttt{INSPECT NONE}: \textbf{\texttt{flatten\_3d=True} destroys walls/goal geometry} (every $2$D maze slice is vectorised before the network sees it) $\rightarrow$ \textbf{add a CNN encoder}};
    \node[rowbox={(n_r)(sn_r)}] (row_r) {};
    \node[nodebar=BrickRed, xshift=3mm] (n_0) at ([yshift=-1.5mm]row_r.south west) {\texttt{[root\_0, hp]} \texttt{config.yaml} truncated $+$ \texttt{num\_train\_steps}$=10^{8}$, XLA crash, \textbf{no score}};
    \node[sidenote] (sn_0) at ([xshift=3mm]n_0.north east) {step 1 \textbf{post-mortem} \texttt{READ src/config.yaml, src/networks.py}: \texttt{root\_0} has no validate output to \texttt{INSPECT}; the only diffs vs.\ baseline are \textbf{truncated config} and \texttt{num\_train\_steps}$=10^{8}$ -- the CNN idea was never actually tested $\rightarrow$ \textbf{implementation error, not idea failure}};
    \node[rowbox={(n_0)(sn_0)}] (row_0) {};
    \draw[connector] ([xshift=2mm]n_r.south west) |- (n_0.west);
    \node[bestbar=deepteal] (n_1) at ([yshift=-1.5mm]row_0.south west) {\texttt{[root\_1, arch]} restored steps $+$ CNN preserving maze geometry, reward $23.03$};
    \node[sidenote] (sn_1) at ([xshift=3mm]n_1.north east) {\textbf{failure-attribution decision}: keep the CNN hypothesis alive; restore \texttt{num\_train\_steps}$=10$M and the original config};
    \node[rowbox={(n_1)(sn_1)}] (row_1) {};
    \draw[connector] ([xshift=2mm]n_r.south west) |- (n_1.west);
    \node[nodebar=deeppurple, xshift=3mm] (n_10) at ([yshift=-1.5mm]row_1.south west) {\texttt{[root\_1\_0, arch]} deeper conv stack, reward $25.96$};
    \node[sidenote] (sn_10) at ([xshift=3mm]n_10.north east) {step 2 \texttt{INSPECT root\_1}: CNN beats baseline by $+\!7.3$. \textbf{Scientist audit: recurrent memory is ``the next logical leap''} (MetaMaze is partially observable), but a GRU/LSTM requires invasive trainer changes $\rightarrow$ verbalized sampler picks the simpler \textbf{deeper conv stack} as a cheaper check};
    \node[rowbox={(n_10)(sn_10)}] (row_10) {};
    \draw[connector] ([xshift=2mm]n_1.south west) |- (n_10.west);
    \node[nodebar=deeppurple, xshift=3mm] (n_100) at ([yshift=-1.5mm]row_10.south west) {\texttt{[root\_1\_0\_0, optim]} $64$ envs, long rollouts, orthogonal init, reward $31.23$};
    \node[sidenote] (sn_100) at ([xshift=3mm]n_100.north east) {step 3 \texttt{READ src/networks.py}: small-arch tweaks plateauing. Scientist again notes memory is the largest remaining info-loss; verbalized sampler picks \textbf{training scale} (envs/rollouts/init) instead because the trainer changes for it are minimal};
    \node[rowbox={(n_100)(sn_100)}] (row_100) {};
    \draw[connector] ([xshift=2mm]n_10.south west) |- (n_100.west);
    \node[nodebar=deeppurple, xshift=3mm] (n_1000) at ([yshift=-1.5mm]row_100.south west) {\texttt{[root\_1\_0\_0\_0, arch]} CNN $+$ scale combo, reward $31.23$ (no gain)};
    \node[sidenote] (sn_1000) at ([xshift=3mm]n_1000.north east) {step 4 \texttt{INSPECT root\_1}: the earlier CNN ran at default scale; MLP$+$scale reached $31.2$, so a \textbf{CNN with the same scale h-params} is the obvious untested combination};
    \node[rowbox={(n_1000)(sn_1000)}] (row_1000) {};
    \draw[connector] ([xshift=2mm]n_100.south west) |- (n_1000.west);
    \node[bestbar=deepteal] (n_1001) at ([yshift=-1.5mm]row_1000.south west) {\texttt{[root\_1\_0\_0\_1, data]} richer egocentric obs, \textbf{reward $48.57$} \;{\large$\bigstar$}};
    \node[sidenote] (sn_1001) at ([xshift=3mm]n_1001.north east) {step 5 \texttt{READ src/networks.py}: CNN$+$scale gave no extra gain -- the spatial encoder isn't the bottleneck either. Single-frame local view confirms \textbf{partial observability is the structural bottleneck} $\rightarrow$ \textbf{enrich the observation} (still no LSTM/GRU)};
    \node[rowbox={(n_1001)(sn_1001)}] (row_1001) {};
    \draw[connector] ([xshift=2mm]n_100.south west) |- (n_1001.west);

    \node[sectionhdr] (h2) at ([yshift=-5mm, xshift=-21mm]row_1001.south west) {AIRA \;-\; 13 nodes, depth 6, best reward $=$ \textbf{48.04}. Proposes LSTM at depth 1, then abandons it.};
    \node[nodebar=deepblue] (a_r) at ([yshift=-2mm]h2.south west) {\texttt{[root]} PPO baseline, reward $15.73$};
    \node[sidenote] (asn_r) at ([xshift=3mm]a_r.north east) {UCB selection only -- no scientist audit; next node maximises UCB};
    \node[rowbox={(a_r)(asn_r)}] (arow_r) {};
    \node[bestbar=deepteal, xshift=3mm] (a_0) at ([yshift=-1.5mm]arow_r.south west) {\texttt{[root\_0]} \textbf{PPO $+$ LSTM/GRU recurrent policy}, \textbf{reward $48.04$} \;{\large$\bigstar$}};
    \node[sidenote] (asn_0) at ([xshift=3mm]a_0.north east) {best of the run at depth $1$: the textbook recurrent solution works on first try};
    \node[rowbox={(a_0)(asn_0)}] (arow_0) {};
    \draw[connector] ([xshift=2mm]a_r.south west) |- (a_0.west);
    \node[nodebar=deeppurple] (a_1) at ([yshift=-1.5mm]arow_0.south west) {\texttt{[root\_1]} Random Network Distillation (RND) intrinsic reward, reward $28.85$};
    \node[sidenote] (asn_1) at ([xshift=3mm]a_1.north east) {orthogonal idea for sparse reward};
    \node[rowbox={(a_1)(asn_1)}] (arow_1) {};
    \draw[connector] ([xshift=2mm]a_r.south west) |- (a_1.west);
    \node[nodebar=deeppurple] (a_2) at ([yshift=-1.5mm]arow_1.south west) {\texttt{[root\_2]} curriculum learning, no score (failed)};
    \node[sidenote] (asn_2) at ([xshift=3mm]a_2.north east) {};
    \node[rowbox={(a_2)(asn_2)}] (arow_2) {};
    \draw[connector] ([xshift=2mm]a_r.south west) |- (a_2.west);
    \node[nodebar=BrickRed, xshift=3mm] (a_00) at ([yshift=-1.5mm]arow_2.south west) {\texttt{[root\_0\_0]} LSTM $+$ RND/ICM stack, \textbf{no score} (training failed)};
    \node[sidenote] (asn_00) at ([xshift=3mm]a_00.north east) {AIRA does not refine LSTM itself; it adds an orthogonal exploration trick on top, which destabilises training};
    \node[rowbox={(a_00)(asn_00)}] (arow_00) {};
    \draw[connector] ([xshift=2mm]a_0.south west) |- (a_00.west);
    \node[nodebar=BrickRed, xshift=3mm] (a_01) at ([yshift=-1.5mm]arow_00.south west) {\texttt{[root\_0\_1]} LSTM $+$ invalid-action masking, reward $19.60$ \;\textbf{regression}};
    \node[sidenote] (asn_01) at ([xshift=3mm]a_01.north east) {same pattern -- LSTM combined with a second change destabilises (UCB does \emph{not} return to bare \texttt{root\_0}'s $48.04$ family)};
    \node[rowbox={(a_01)(asn_01)}] (arow_01) {};
    \draw[connector] ([xshift=2mm]a_0.south west) |- (a_01.west);
    \node[nodebar=gray, xshift=3mm] (a_ch) at ([yshift=-1.5mm]arow_01.south west) {$5$ further LSTM/GRU attempts deeper in the tree (\texttt{root\_1\_0}, \texttt{root\_0\_1\_0\_0}, \texttt{root\_1\_0\_0\_0}, \ldots): all stacked with RND/ICM/larger nets, all score $17.5$--$22.3$, never returns to bare LSTM};
    \node[sidenote] (asn_ch) at ([xshift=3mm]a_ch.north east) {AIRA proposes LSTM repeatedly but \emph{never} ablates back to the bare $48.04$ configuration};
    \node[rowbox={(a_ch)(asn_ch)}] (arow_ch) {};
    \draw[connector] ([xshift=2mm]a_0.south west) |- (a_ch.west);
    \end{tikzpicture}%
    }
    \caption{MetaMaze search trees. \shortname{} (top, $7$ nodes) recovers
    from a config-truncation crash via failure attribution, then climbs
    $15.73 \rightarrow 48.57$ on a pure-MLP path -- the scientist explicitly
    identifies recurrent memory as the missing axis at steps 2, 3 and 5 but
    the verbalized sampler always selects a cheaper alternative. AIRA (bottom,
    $13$ nodes) proposes a PPO$+$LSTM recurrent policy at \texttt{root\_0}
    and reaches $\mathbf{48.04}$ on first try, but UCB then deepens elsewhere
    and never returns to refine the bare LSTM; $5$ further LSTM attempts
    deeper in the tree all combine LSTM with RND/ICM/larger nets and
    destabilise at $17.5$--$22.3$. Both methods peak near $48$ for opposite
    reasons; the full LSTM gain ($48 \rightarrow \approx\!90$) is recovered
    by test-time training (Table~\ref{tab:ttt_mlgym}, MetaMaze:
    $30.4 \rightarrow \mathbf{53.0}$).}
    \label{fig:metamaze_tree}
\end{figure}
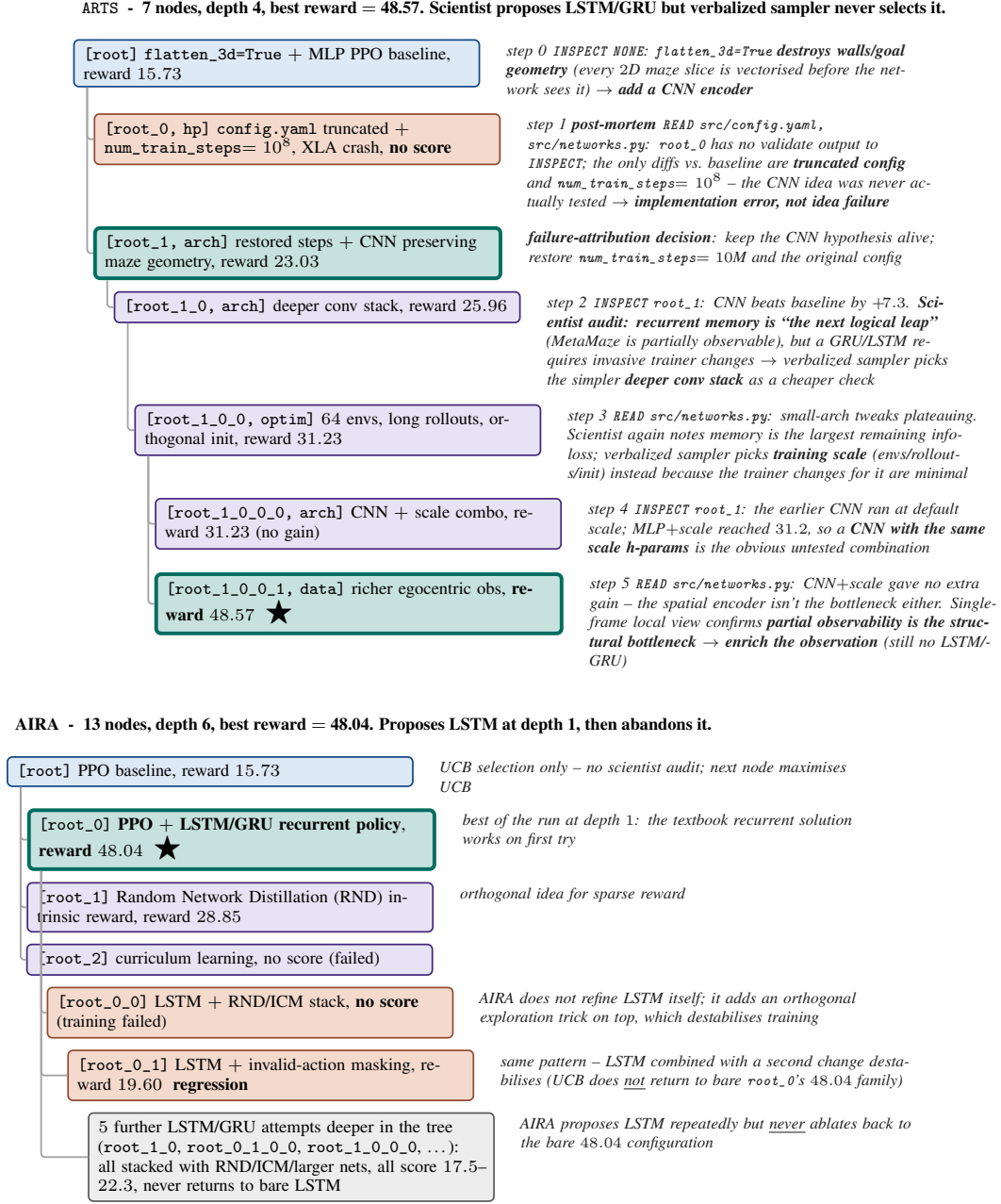

We contrast \shortname{} and AIRA on MLGym MetaMaze (mean evaluation reward,
higher is better; ceiling $\approx\!100$). Both methods reach nearly
identical peaks ($48.57$ for \shortname{}, $48.04$ for AIRA), but for
\emph{opposite} reasons. MetaMaze is partially observable, so a recurrent
policy (LSTM/GRU) is the textbook solution. \shortname{}'s scientist
repeatedly identifies recurrent memory as ``the next logical leap'' in its
reasoning but the verbalized sampler keeps picking simpler alternatives
(deeper conv, training scale, richer observation), so \emph{no LSTM/GRU
node is ever actually run}. AIRA proposes LSTM at \texttt{root\_0} and
reaches the run's best ($48.04$) on its first try, then deepens the wrong
branch, abandons \texttt{root\_0}, and stacks RND/ICM on top of $5$ further
LSTM attempts which all destabilise at $17.5$--$22.3$.

\section{MountainCar Search Trajectory: \shortname{} versus AIRA}
\label{app:mountaincar_tree}

\shortname{} and AIRA on MLGym MountainCarContinuous (mean reward, higher is
better; ceiling $\approx\!99$). Peak scores are similar ($\sim\!95$) but
\shortname{} reaches reward $95.73$ in $4$ steps at depth $4$; AIRA reaches
$94.82$ at depth $13$ after $39$ expansions ($\sim\!10\!\times$ the compute),
with one root draft diverging to $-9{,}708.99$. The MountainCar margin in
Table~\ref{tab:ttt_mlgym} is primarily a \emph{robustness} gap (mean
ARTS $95.73_{\pm.69}$ vs.\ mean AIRA $80.82_{\pm 11.75}$). Italic margin
notes give the scientist's tool call and the reasoning that drove the next
experiment.

\begin{figure}[H]
    \centering
    \resizebox{\linewidth}{!}{%
    \begin{tikzpicture}[
        font=\scriptsize,
        nodebar/.style={
            draw=#1!70!black,
            fill=#1!14,
            rounded corners=2pt,
            line width=0.65pt,
            align=left,
            anchor=north west,
            text width=5.5cm,
            inner xsep=4pt,
            inner ysep=3pt
        },
        bestbar/.style={
            draw=#1!90!black,
            fill=#1!22,
            rounded corners=2pt,
            line width=1.5pt,
            align=left,
            anchor=north west,
            text width=5.5cm,
            inner xsep=4pt,
            inner ysep=3pt
        },
        sidenote/.style={
            align=left,
            anchor=north west,
            text width=5.8cm,
            inner xsep=2pt,
            inner ysep=2pt,
            font=\fontsize{6.5pt}{7.8pt}\selectfont\itshape,
            text=gray!25!black
        },
        rowbox/.style={fit=#1, draw=none, inner sep=0pt},
        connector/.style={draw=gray!72, line width=0.7pt, rounded corners=1.5pt},
        sectionhdr/.style={anchor=north west, font=\scriptsize\bfseries}
    ]
    \node[sectionhdr] (h1) at (0,0) {\shortname{} \;-\; 9 nodes, depth 4, best reward $=$ \textbf{95.73}};
    \node[nodebar=deepblue] (n_r) at ([yshift=-2mm]h1.south west) {\texttt{[root]} PPO baseline (default h-params), reward $33.79$};
    \node[sidenote] (sn_r) at ([xshift=3mm]n_r.north east) {step 0 \texttt{READ src/networks.py, policy.py, train.py}: the $2$-dim obs (pos, vel) throws nothing away; the real bottleneck is the \textbf{sparse reward} ($+100$ only at $0.45$\,m) $\rightarrow$ \textbf{tune PPO h-params} (entropy$\uparrow$ for exploration, $\gamma\!\uparrow$ for credit assignment)};
    \node[rowbox={(n_r)(sn_r)}] (row_r) {};
    \node[nodebar=deeppurple, xshift=3mm] (n_0) at ([yshift=-1.5mm]row_r.south west) {\texttt{[root\_0, hp]} tuned PPO (entropy, $\gamma$, $\lambda_{GAE}$, lr), reward $94.73$ ($+\!60.94$)};
    \node[sidenote] (sn_0) at ([xshift=3mm]n_0.north east) {step 1 \texttt{INSPECT root\_0}: $+60$ from h-params alone; \textbf{actor uses a fixed-scalar action $\sigma$}, which wastes exploration once the policy converges $\rightarrow$ \textbf{make $\sigma$ a function of state}};
    \node[rowbox={(n_0)(sn_0)}] (row_0) {};
    \draw[connector] ([xshift=2mm]n_r.south west) |- (n_0.west);
    \node[nodebar=deeppurple, xshift=3mm] (n_00) at ([yshift=-1.5mm]row_0.south west) {\texttt{[root\_0\_0, arch]} state-dependent action $\sigma$ (head outputs $\sigma$), reward $94.83$};
    \node[sidenote] (sn_00) at ([xshift=3mm]n_00.north east) {step 2 \textbf{exploit} \texttt{INSPECT root\_0\_0}: near peak ($94.83$), but \textbf{no regularization tried}; advantage is unnormalised, value loss unclipped $\rightarrow$ \textbf{standard PPO finish: adv-norm + value-clip + tuned entropy}};
    \node[rowbox={(n_00)(sn_00)}] (row_00) {};
    \draw[connector] ([xshift=2mm]n_0.south west) |- (n_00.west);
    \node[bestbar=deepteal, xshift=3mm] (n_000) at ([yshift=-1.5mm]row_00.south west) {\texttt{[root\_0\_0\_0, reg]} adv-norm $+$ value-clip $+$ entropy, \textbf{reward $95.73$} \;{\large$\bigstar$}};
    \node[sidenote] (sn_000) at ([xshift=3mm]n_000.north east) {steps 3-6 \texttt{INSPECT root\_0\_0\_0}: \textbf{$95.73 \approx$ ceiling}; probe the remaining axes (data, arch, loss, optim) around the winner to confirm no further easy gain};
    \node[rowbox={(n_000)(sn_000)}] (row_000) {};
    \draw[connector] ([xshift=2mm]n_00.south west) |- (n_000.west);
    \node[nodebar=gray, xshift=3mm] (n_ref) at ([yshift=-1.5mm]row_000.south west) {\texttt{[root\_0\_0\_0\_\{0..4\}]} richer obs ($88.4$), wider net ($28.8$, collapsed), \texttt{tanh} ($87.5$), $4$ hidden layers ($95.7$), obs proc ($87.4$)};
    \node[sidenote] (sn_ref) at ([xshift=3mm]n_ref.north east) {only \texttt{..\_0\_0\_0\_3} ($4$ hidden layers, $95.7$) stays close; \textbf{the regularization regime is the dominant lever}};
    \node[rowbox={(n_ref)(sn_ref)}] (row_ref) {};
    \draw[connector] ([xshift=2mm]n_000.south west) |- (n_ref.west);

    \node[sectionhdr] (h2) at ([yshift=-5mm, xshift=-12mm]row_ref.south west) {AIRA \;-\; 39 nodes, depth 13, best reward $=$ \textbf{94.82}};
    \node[nodebar=deepblue] (a_r) at ([yshift=-2mm]h2.south west) {\texttt{[root]} PPO baseline, reward $33.79$};
    \node[sidenote] (asn_r) at ([xshift=3mm]a_r.north east) {UCB selection only -- no scientist; the next node is whichever leaf maximises UCB};
    \node[rowbox={(a_r)(asn_r)}] (arow_r) {};
    \node[nodebar=deeppurple, xshift=3mm] (a_0) at ([yshift=-1.5mm]arow_r.south west) {\texttt{[root\_0]} draft $1$: actor-critic MLP, reward $27.06$};
    \node[sidenote] (asn_0) at ([xshift=3mm]a_0.north east) {};
    \node[rowbox={(a_0)(asn_0)}] (arow_0) {};
    \draw[connector] ([xshift=2mm]a_r.south west) |- (a_0.west);
    \node[nodebar=BrickRed] (a_1) at ([yshift=-1.5mm]arow_0.south west) {\texttt{[root\_1]} draft $2$: alt parameterisation, reward $\mathbf{-10{,}779}$};
    \node[sidenote] (asn_1) at ([xshift=3mm]a_1.north east) {catastrophic divergence};
    \node[rowbox={(a_1)(asn_1)}] (arow_1) {};
    \draw[connector] ([xshift=2mm]a_r.south west) |- (a_1.west);
    \node[nodebar=deeppurple] (a_2) at ([yshift=-1.5mm]arow_1.south west) {\texttt{[root\_2]} draft $3$: tuned PPO, reward $38.16$};
    \node[sidenote] (asn_2) at ([xshift=3mm]a_2.north east) {UCB picks this; deepens for $17$ levels};
    \node[rowbox={(a_2)(asn_2)}] (arow_2) {};
    \draw[connector] ([xshift=2mm]a_r.south west) |- (a_2.west);
    \node[nodebar=gray, xshift=3mm] (a_ch) at ([yshift=-1.5mm]arow_2.south west) {\texttt{root\_2\_0}$\rightarrow\cdots$: depth-$13$ chain of PPO variants, deepest at depth $18$};
    \node[sidenote] (asn_ch) at ([xshift=3mm]a_ch.north east) {};
    \node[rowbox={(a_ch)(asn_ch)}] (arow_ch) {};
    \draw[connector] ([xshift=2mm]a_2.south west) |- (a_ch.west);
    \node[bestbar=deepteal, xshift=3mm] (a_w) at ([yshift=-1.5mm]arow_ch.south west) {depth $13$: PPO with regularization, \textbf{reward $94.82$} \;{\large$\bigstar$}};
    \node[sidenote] (asn_w) at ([xshift=3mm]a_w.north east) {comparable peak, $\sim\!10\!\times$ the budget};
    \node[rowbox={(a_w)(asn_w)}] (arow_w) {};
    \draw[connector] ([xshift=2mm]a_ch.south west) |- (a_w.west);
    \end{tikzpicture}%
    }
    \caption{MountainCar search trees. \shortname{} (top) makes three
    sequential moves --- tune PPO h-params, add state-dependent action
    $\sigma$, add regularization --- and reaches reward $\mathbf{95.73}$ at
    depth $4$. AIRA (bottom) drafts three roots; one diverges
    catastrophically, UCB commits to the $38.16$ draft, chains $17$ further
    levels to $94.82$. The table margin reflects baseline runs that
    frequently diverge.}
    \label{fig:mountaincar_tree}
\end{figure}

\section{LLM Use}
\label{llm}
We use LLMs as coding assistance to write code for this project and we manually verify the code at all times. It was also used for light refinement of the paper like making text terse.


\section{Qualitative Analysis of Search Efficiency}
\label{app:qualitative_search_efficiency}

This section studies whether \shortname{} searches more efficiently than the baselines. We use the logs to ask the following questions.
\begin{enumerate}[leftmargin=*,noitemsep,topsep=2pt]
    \item Can \shortname{} separate a promising hypothesis from a bad execution?
    \item Does auditing identify the root cause of failed or regressed experiments?
    \item Does the scientist reason from logs rather than only follow final scores?
    \item Does reasoning-based parent selection spend budget better than heuristic selection?
    \item Can the scientist propose directions that are not already in the tree?
    \item Does verbalized sampling preserve enough diversity?
    \item How do baselines fail differently under the same budget?
\end{enumerate}
We answer each question separately below.

\subsection{Can \shortname{} separate a promising hypothesis from a bad execution?}

\textbf{Answer: yes.} The logs show that \shortname{} often preserves a promising hypothesis after a flawed implementation. On HMS Brain Activity, the task is to predict seizure-related activity from EEG spectrograms, and a central difficulty is how to aggregate multiple windows for the same patient. Using more spectrogram windows is promising because the single-window baseline discards much of the labeled signal. The first all-window attempt regressed from $0.963$ to $1.237$ KL. \shortname{} did not abandon the idea. It kept the same broad hypothesis and changed the implementation. On Vesuvius, where the task is to segment hidden ink from 3D scroll scans, an EMA regularization attempt returned $0.000$ $F_{0.5}$ because the model predicted an all-zero mask. \shortname{} treated this as a broken EMA implementation, not as evidence that EMA was a bad idea. On FineWeb language modeling, a longer-training attempt timed out after reducing the per-device batch size. \shortname{} kept the longer-training hypothesis and later moved validation loss from $4.67$ to $4.29$ and then to $3.97$.

\subsection{Does auditing identify the root cause of failed or regressed experiments?}

\textbf{Answer: yes.} The audit is useful when the score alone is ambiguous. In HMS, the failed all-window node looked like a bad direction from the metric alone. The audit instead found two concrete implementation errors: duplicated \texttt{eeg\_id}s in the split and averaging probabilities after softmax instead of aggregating logits before softmax. In Vesuvius, the all-zero prediction under EMA was traced to how the EMA model was maintained and evaluated. The next proposal maintained shadow weights after each optimizer step and used the EMA model only for validation and inference. In FineWeb, the timeout was diagnosed as a throughput problem, not as evidence against training longer. These diagnoses are specific enough to change the next experiment.

\subsection{Does the scientist reason from logs rather than only follow final scores?}

\textbf{Answer: yes.} Several decisions depend on logged failure modes rather than the final score alone. On FineWeb, the scientist inspected the training curves and concluded that the model was under-trained, so it extended training instead of only sweeping batch size or attention heads. On APTOS, a naive resolution increase to $384^2$ ran successfully but reduced QWK from $0.932$ to $0.906$. \shortname{} did not keep increasing resolution. It reasoned that resolution would only help with better cropping, augmentation, or regularization. On HMS, replacing ResNet18 with ResNet34 under the same pipeline worsened KL from $0.541$ to $0.627$. Since the run was valid, the scientist treated this as evidence against simply scaling the backbone. These are log-level inferences, not score-only choices.

\subsection{Does reasoning-based parent selection spend budget better than heuristic selection?}

\textbf{Answer: yes, by both qualitative evidence and automatic proxies.} A score alone cannot tell whether a node failed because the idea was weak or because the implementation was bad. \shortname{} uses inspection, memory, and reasoning to choose which parent deserves another child. This lets it return to a promising branch after an implementation failure, while avoiding branches that ran cleanly but regressed. We also compute a rough proxy for node-selection quality. We define a \emph{valid-but-regressing} node as a node that produces a valid score, does not have a fatal execution marker, and is worse than its parent. Such nodes are an imperfect proxy for bad ideas with good execution. On selected tasks with known metric direction, \shortname{} expanded only $1.4\%$ of valid-but-regressing nodes, compared with $92.7\%$ for AIRA and $64.2\%$ for MLEvolve. The mean number of descendants below such nodes was $0.76$ for \shortname{}, $30.3$ for AIRA, and $6.23$ for MLEvolve. Among internal nodes, the fraction that eventually led to an improving descendant was $80.4\%$ for \shortname{}, compared with $75.8\%$ for AIRA and $64.1\%$ for MLEvolve. These numbers are noisy because execution status is not uniformly logged across methods, but they support the qualitative pattern that \shortname{} spends less budget on cleanly executed but unpromising branches.

\subsection{Can the scientist propose directions not already in the tree?}

\textbf{Answer: yes.} The scientist is not restricted to mutating the current best code. It can read the tree, inspect logs, consult memory, and propose a new hypothesis before the executor writes code. On HMS, after a strong ResNet18 branch, the scientist did not simply continue local tuning. It listed three directions: combine full-window coverage with the strong backbone, try a larger pretrained architecture, and change the loss for class imbalance. These candidates were assigned probabilities of $0.50$, $0.30$, and $0.20$, and the sampled experiment combined the relevant branches. This is evidence that the scientist can propose directions not already represented by the current best lineage.

\subsection{Does verbalized sampling preserve enough diversity?}

\textbf{Answer: yes.} Verbalized sampling forces the scientist to list multiple candidate hypotheses with explicit probabilities before the system samples one. The probabilities represent the scientist's current belief about which hypotheses are worth trying after inspecting the tree, logs, and memory. This makes the proposal distribution visible and keeps the search from collapsing to a single local edit. We measure rough strategy coverage with keyword labels over node hypotheses. \shortname{} visits $4.11$ unique axes per run on average, compared with $2.20$ for AIRA and $2.43$ for MLEvolve. The axis-distribution entropy is $1.16$ for \shortname{}, $0.43$ for AIRA, and $0.58$ for MLEvolve. These numbers support the claim that verbalized sampling increases diversity. However, diversity alone is not enough. The stronger finding is that \shortname{} combines diversity with inspection: it opens new axes when the audit or memory says an axis is missing, and it returns to old parents when a promising idea failed for an execution reason.

\subsection{How do baselines fail differently under the same budget?}

\textbf{Answer: the baselines fail in method-specific ways.} Linear search is brittle because it continues the last node and has no explicit mechanism to return to an older promising parent after a bad implementation. AIRA can sometimes describe that a failure is likely a bug, but the diagnosis is less reliably coupled to the next parent choice. In HMS, AIRA expanded an unstable spectrogram-CNN branch whose score had already degraded to $4.08$ KL; the next child worsened further to $9.43$ KL. On FineWeb, AIRA produced root-level attempts without a valid submission, including a CUDA out-of-memory failure during validation, and did not recover a scored branch from that failure. MLEvolve explores through mutation and crossover, which can be useful for compact programs, but its reflections are sometimes weakly grounded in the code that actually ran. In MetaMaze, for example, it sometimes attributes performance to recurrent LSTM/GRU memory even when the logged branch still uses a flattened feedforward representation. These failures are not just lower scores. They show weaker coupling between evidence, parent selection, and the next executable experiment.

\section{Failure Attribution: Detailed Examples}
\label{app:failure_attribution}

This appendix expands the main-text Finding~1 with full per-node
trajectories. Each example contrasts what a score-only policy would
read from the regression with what \shortname{}'s \emph{idea-wrong} /
\emph{implementation-wrong} attribution actually inferred, and what
the next expansion did as a consequence.

\subsection{HMS Brain Activity: full trajectory}
\label{app:fa_hms}

HMS Harmful Brain Activity Classification (MLE-bench; KL divergence,
\emph{lower} is better; uniform-prediction baseline KL $1.462$). The
trajectory below is from the \shortname{} run
\texttt{llmg\_C\_vs\_C\_v3} reported in Table~\ref{tab:main}. The full
tree with scientist tool calls is in
Fig.~\ref{fig:hms_tree}~(Appendix~\ref{app:hms_search_tree}).

\textbf{Initial drafts (depth 1).} \shortname{} drafts five families in
parallel. The strongest is a pretrained ResNet-50 over 3-channel
log-mel spectrograms (\texttt{root\_2}, KL $0.557$); a 1-channel
ResNet-18 baseline (\texttt{root\_0}, $0.623$) and a data-axis
variant (\texttt{root\_1}, $0.562$) are sibling families.

\textbf{Two consecutive regressions (depth 2).} The two children of
the strong ResNet-50 family both regress:
\texttt{root\_2\_0} swaps the backbone to EfficientNet-B0 and lands at
KL $0.702$ (a $+0.145$ regression);
\texttt{root\_2\_1} keeps ResNet-50 and adds SpecAugment ($0$--$2$
frequency masks $\le 12$ bins, $0$--$2$ time masks $\le 20\%$ of time
steps) and lands at KL $0.565$ (a small $+0.008$ regression). A
score-only policy reads two regressions in a row as evidence that the
spectrogram-CNN family is exhausted.

\textbf{Failure-attribution decision.} \shortname{} re-reads the
training logs of both regressed nodes. The validation KL is still
decreasing at the last epoch in both runs --- SpecAugment in
particular makes the optimisation slower because it injects
augmentation noise. The scientist classifies the regression as
\textbf{implementation-wrong} (under-trained under heavy augmentation,
not a failure of the spectrogram representation) and retains the
ResNet-50 + spectrogram family.

\textbf{Recovery and refinement (depth 3+).} The next expansions stay
on the ResNet-50 + spectrogram axis and improve KL monotonically:
\texttt{root\_2\_2} adds Mixup with Beta$(0.4, 0.4)$ to ResNet-50
(KL $0.527$, the first improvement over the parent);
\texttt{root\_2\_2\_2} stacks Mixup and SpecAugment together
($0.519$);
\texttt{root\_2\_2\_2\_0} adds per-channel normalisation with
dataset-level statistics
($\mu, \sigma$ computed across the $10{,}024$ training spectrograms:
$\mu_{C_0}\!=\!1.10$, $\mu_{C_1}\!=\!0.74$, $\mu_{C_2}$ similar) to
reach $0.516$. Continued refinement on the same family ultimately
reaches \textbf{KL $\mathbf{0.467}$} --- the best on the task.

\textbf{Contrast with AIRA on the same task.} AIRA's first draft also
finds a strong spectrogram-CNN family at KL $0.513$. Its next two
expansions, however, pivot to entirely different pipelines: a
hand-crafted feature pipeline using \texttt{scipy.signal} with
\texttt{skew}/\texttt{kurtosis} statistics, and a derivative-stacked
EfficientNet over the same spectrograms. The children of these pivots
diverge catastrophically --- \texttt{root\_2\_0} reaches KL $4.08$
(roughly $3\times$ worse than baseline) and \texttt{root\_2\_0\_0}
reaches KL $9.43$ --- and AIRA never returns to the original
$0.513$ family. The cost of conflating ``why'' with ``that'' is
visible here: a working family abandoned after one node is a working
family the search never refines.

\subsection{APTOS Diabetic Retinopathy: pretrained-feature collapse}
\label{app:fa_aptos}

APTOS 2019 (MLE-bench; quadratic-weighted kappa, higher is better).

\textbf{The collapsed first attempt.} The initial draft is a
ConvNeXt-Base backbone, fully fine-tuned end-to-end at learning rate
$10^{-3}$. The training loss decreases over the first epoch but the
held-out QWK is exactly $\mathbf{0.000}$: the high learning rate
erases the pretrained ImageNet features within the first epoch and
the model converges to a constant prediction.

\textbf{Score-only attribution.} A score-only policy reads the
$0.000$ as evidence that \emph{ConvNeXt is the wrong backbone for this
task} and pivots to a different architecture family (typical baseline
responses are ResNet-50, EfficientNet, or a custom shallow CNN).

\textbf{\shortname{}'s attribution.} The scientist inspects the
training log, observes that the train loss decreases monotonically
while the validation QWK is pinned at $0.000$, and reads this as
collapse to a degenerate constant prediction --- consistent with the
learning rate being too large to preserve the pretrained features.
The regression is classified as
\textbf{implementation-wrong (learning rate schedule)} and the
ConvNeXt-Base backbone is retained.

\textbf{Corrected execution.} The next expansion edits the training
script to:
(i) train the classification head only for one epoch with the
backbone frozen (linear-probe warmup); then
(ii) unfreeze the backbone and continue at learning rate
$3\!\times\!10^{-5}$ with the same optimiser.
The corrected run reaches QWK $\mathbf{0.92}$.

\subsection{Vesuvius Ink Detection: a broken validation split}
\label{app:fa_vesuvius}

Vesuvius Ink Detection (MLE-bench; $F_{0.5}$, higher is better).

\textbf{A misleadingly low score.} The first DeepLabV3-ResNet50 draft
uses a 3-channel input ($z\!-\!1, z, z\!+\!1$) and reaches
$F_{0.5}=0.151$. A subsequent expansion widens the input to 11
channels ($z\!-\!5$ to $z\!+\!5$, intended to capture more z-axis
context) and lands at $F_{0.5}=0.165$ --- barely above the
3-slice version.

\textbf{Score-only attribution.} A score-only policy reads the
$0.165$ as evidence that the wider receptive field does not help,
prunes the wide-input branch, and pivots to a different architecture
(this is what AIRA does on the same task --- see
Fig.~\ref{fig:vesuvius_tree} --- chaining 3D U-Net, 2.5D U-Net, CRNN,
PointNet, sparse-3D conv, Video-ViT, and Swin-UNETR down a single
chain, each abandoned after one node).

\textbf{\shortname{}'s attribution.} The scientist re-reads the
training log line by line. The log contains the line
\texttt{DATA: train$=$4000 val$=$0}: the validation split was set to
zero by the training script, so \texttt{val\_metric} is constant at
zero throughout training and the reported $F_{0.5}=0.165$ is computed
on an unrepresentative held-out shard. The regression is classified
as \textbf{implementation-wrong (broken validation split)} rather than
as a failure of the wide-input idea.

\textbf{Corrected execution.} The next expansion retains the
DeepLabV3 + wide-slice family with the validation split repaired,
then adds the second scroll fragment to the training data
(\texttt{root\_3}, $F_{0.5}=0.479$ --- a large jump), and finally
adds tile/flip augmentation on top of fragment-2 (\texttt{root\_3\_1},
$F_{0.5}=\mathbf{0.575}$), the best on the task.

\subsection{Why the same pattern shows up across tasks}
\label{app:fa_summary}

The three examples are not isolated. They share a common structure:
a hypothesis that is sound \emph{as a research direction} surfaces a
low score because of an execution-side artefact (an over-aggressive
learning rate, an under-trained model under augmentation, a broken
validation split). A score-only policy reads the low score as
evidence against the hypothesis and pivots; the working family is
then either lost entirely (HMS, Vesuvius under AIRA) or rediscovered
many nodes later in a degraded form (Vesuvius: AIRA only returns to
the 2.5D-pretrained variant at depth $13$, $F_{0.5}=0.510$).
\shortname{}'s requirement to label every regression as
\emph{idea-wrong} or \emph{implementation-wrong} forces the scientist
to inspect the code and the log before reassigning the budget, which
in each case here points to a specific, fixable execution issue
rather than to abandonment.

\newpage
\section{Extended Related Work}
\label{app:extended_related_work}

\para{Autonomous scientific discovery.}
Recent systems use language models as proposal engines for scientific discovery. FunSearch~\citep{funsearch} combines an LLM with an evaluator in an evolutionary loop and discovers new constructions in extremal combinatorics. AlphaEvolve~\citep{alphaevolve} uses Gemini as a code mutation engine and reports improvements in algorithmic domains such as matrix multiplication. CosScientist~\citep{coscientist} and Google's AI Co-Scientist~\citep{google_ai_coscientist} study hypothesis generation and planning in wet-lab settings. The Sakana AI Scientist~\citep{sakana_ai_scientist} automates parts of the ML research pipeline, including hypothesis generation, code writing, execution, and write-up. These systems differ from our setting in both domain and search structure. We study ML research tasks where each candidate requires writing and running training code, and where the main question is how to choose which prior node to expand.

\para{Linear agents.}
MLGym~\citep{mlgym} and MLAgentBench~\citep{mlagentbench} evaluate agents that interact with ML workspaces through sequential actions. AutoResearch~\citep{autoresearch} adds planning, but still follows a single trajectory. This structure is computationally simple and easy to deploy, but it commits strongly to early decisions. Reasoning models can backtrack within a chain of thought~\citep{deepseekr1, qin2025backtrack}, but the revision happens inside the current context window. In long ML runs, the agent accumulates code, logs, validation scores, and failure traces. Prior work on long-context models shows that information retrieval and reasoning degrade as relevant evidence is pushed deeper into context~\citep{liu2023lost, hsieh2024ruler}. This makes linear search fragile when the useful direction is an earlier hypothesis that failed for implementation reasons.

\para{Tree search.}
Tree-based agents keep an explicit search tree and choose a node to expand. Greedy variants such as AIDE~\citep{aide} expand the best-scoring node, while MCTS variants such as AIRA~\citep{aira} and AI Scientist v2~\citep{sakana_ai_scientist} use UCT-style selection. These methods import a useful abstraction from games, but ML research does not have cheap or reliable rollouts. A node score is a noisy mixture of hypothesis quality, implementation correctness, data handling, training stability, and evaluation details. A valid idea can receive a poor score because of a coding error, and a weak idea can receive a high score because the implementation is easier. Score-based backpropagation therefore has the wrong primitive for deciding which scientific direction deserves more attempts.

\para{Evolutionary search.}
Evolutionary and population-based systems such as FunSearch~\citep{funsearch}, AlphaEvolve~\citep{alphaevolve}, OpenEvolve~\citep{openevolve}, and MLEvolve~\citep{mlevolve} maintain a population of programs and apply mutation or crossover. This is effective when the artifact being searched is short and compositional. ML training pipelines are less modular. Data loading, model architecture, losses, augmentations, optimizers, and validation logic interact in tightly coupled ways. Copying a block from one branch into another can break hidden assumptions rather than combine useful ideas. Evolutionary search also spends many evaluations on program variants, which is expensive when each node can require minutes to hours of GPU time.

\para{LLM-guided search.}
Several works use language models to replace or augment hand-designed search heuristics. Verbalized sampling~\citep{verbalized_sampling} asks the model to express a distribution over candidates, which can reduce collapse to a single likely continuation. We use this idea when the scientist proposes hypotheses, because greedy continuation often produces small variants of the current best branch. The key difference in \shortname{} is that the scientist also selects the parent node. It can inspect logs and code from prior attempts and distinguish a failed implementation from a failed hypothesis.

\para{Test-time training.}
Test-time training was introduced as a way to adapt a model on each test instance~\citep{ttt}. Parameter-efficient variants use adapters such as LoRA to update only a small set of weights~\citep{akyurek2025the}. TTT-Discover~\citep{ttt-discover} and execution-grounded agent training~\citep{si2026execution} apply on-policy optimization to research agents, using rollout outcomes as the learning signal. Our setting differs in the structure of the policy. We train a tree-structured scientist that chooses what to inspect, which parent to expand, and which hypothesis to test. This lets the model absorb useful search experience into its weights while still using execution feedback from the current task.

\newpage
\section{Additional Experimental Details}
\label{app:experimental_details}

\para{Execution environment.}
All runs execute inside Apptainer containers with the same executor sandbox, action set, and validation interface. We restore the evaluation script before every validation. We added this guard because agents sometimes attempted to modify evaluation code despite instructions not to do so. Restoring the script prevents metric hacking and keeps the comparison focused on legitimate changes to the submission or training code.

\para{Models and prompts.}
For the main \shortname{} experiments, the scientist is OpenAI o3~\citep{openai_2025} and the executor is Gemini 3 Flash~\citep{gemini3}. For baselines that do not separate scientist and executor roles, we use Gemini 3 Flash for both reasoning and coding. We also report ablations that swap the scientist and executor models. The scientist and executor prompts are given in Appendix~\ref{app:scientist_prompts}.

\para{\shortname{} settings.}
Each run receives an 8-hour wall-clock budget. \shortname{} uses an audit length $R=3$ scientist calls. During proposal, the scientist enumerates $K=5$ candidate hypotheses using verbalized sampling, as described in \S\ref{subsec:scientist}. One candidate is sampled and passed to the executor, which implements and validates the experiment in the sandbox.

\para{TTT settings.}
For test-time training, we use Qwen3-4B-Instruct~\citep{qwen3} as the scientist. TTT uses LoRA adapters of rank 32 with $\alpha=64$ and dropout 0 on all attention and MLP linear projections. We train with AdamW using learning rate $1\!\times\!10^{-5}$, constant schedule, $\beta=(0.9,0.999)$, weight decay 0.01, and gradient clipping at 1.0. Each GRPO step samples $N=8$ rollouts per group and applies one inner update with KL penalty $\beta_{\mathrm{KL}}=0.01$ against the frozen pretrained base. Per-token importance ratios are clipped to $[0.8,1.28]$.

\para{Rollout limits.}
Rollouts use sequence length 8192 with up to 120 scientist turns of 1024 tokens each. Each sampled candidate is executed once, with no best-of-$N$ reranking. LoRA adapters are checkpointed and broadcast to the inference server every 5 GRPO steps. Training is capped at 100 GRPO steps or the 8-hour wall-clock budget, whichever comes first.

\para{Compute.}
All experiments use 40\,GB A100 GPUs. Each inference-only experiment, including Linear, AIRA, MLEvolve, and \shortname{} without TTT, uses one GPU for the executor sandbox, for 8 GPU-hours per run. Each TTT experiment uses three GPUs, one for the LoRA trainer, one for the vLLM inference server hosting Qwen3-4B, and one for the executor sandbox, for 24 GPU-hours per run.

\section{Sources for Human-Best Scores}
\label{app:references_to_human_best}

Table~\ref{tab:human_best_sources} records the provenance of the human-best values reported in Table~\ref{tab:main}. For MLEBench tasks, the value is the top-1 Kaggle leaderboard score for the corresponding competition. For MLGym tasks, the value is the published best-known result for the benchmark task or environment. We include the table to make the normalization constants auditable.

\begin{table}[h]
  \centering
  \setlength{\tabcolsep}{5pt}
  \renewcommand{\arraystretch}{1.05}
  \small
  \begin{tabular}{lcp{0.54\linewidth}}
    \toprule
    Task & Human Best & Source \\
    \midrule
    Titanic & 0.830 & MLGym~\cite{mlgym}; Kaggle task~\cite{kaggle_titanic} \\
    CIFAR-10 & 0.994 & MLGym~\cite{mlgym}; CIFAR-10~\cite{krizhevsky2009learning} \\
    Fashion MNIST & 0.968 & MLGym~\cite{mlgym}; Fashion-MNIST~\cite{xiao2017fashion} \\
    House Price & 0.990 & MLGym~\cite{mlgym}; Ames Housing~\cite{decock2011ames} \\
    MNLI & 92.50 & MLGym~\cite{mlgym}; MultiNLI \\
    Lang.\ Modeling & 3.500 & MLGym~\cite{mlgym}; FineWeb~\cite{penedo2024fineweb} \\
    Battle of Sexes & 1.667 & MLGym task specification~\cite{mlgym} \\
    Prisoner's Dilemma & 3.000 & MLGym task specification~\cite{mlgym} \\
    Blotto & 0.500 & MLGym task specification~\cite{mlgym} \\
    MountainCar & 99.00 & MLGym~\cite{mlgym}; OpenAI Gym~\cite{brockman2016openai} \\
    Breakout & 100.00 & MLGym~\cite{mlgym}; MinAtar~\cite{young2019minatar} \\
    Meta Maze & 52.50 & MLGym task specification~\cite{mlgym} \\
    \midrule
    Spaceship Titanic & 0.828 & MLE-bench/Kaggle leaderboard~\cite{mlebench,kaggle_spaceship_titanic} \\
    Nomad 2018 & 0.051 & MLE-bench/Kaggle leaderboard~\cite{mlebench,sutton2019nomad} \\
    Jigsaw Toxic & 0.989 & MLE-bench/Kaggle leaderboard~\cite{mlebench,kaggle_jigsaw_toxic} \\
    APTOS 2019 & 0.936 & MLE-bench/Kaggle leaderboard~\cite{mlebench,kaggle_aptos2019} \\
    Plant Pathology & 0.984 & MLE-bench/Kaggle leaderboard~\cite{mlebench,thapa2020plantpathology} \\
    Histopathologic Cancer & 1.000 & MLE-bench/Kaggle leaderboard~\cite{mlebench,kaggle_histopathologic_cancer} \\
    Vesuvius Ink Detection & 0.831 & MLE-bench/Kaggle leaderboard~\cite{mlebench,kaggle_vesuvius_ink} \\
    Kuzushiji Recognition & 0.950 & MLE-bench/Kaggle leaderboard\\
    HMS Brain Activity & 0.272 & MLE-bench/Kaggle leaderboard~\cite{mlebench,kaggle_hms_brain_activity} \\
    RSNA Brain Tumor & 0.621 & MLE-bench/Kaggle leaderboard~\cite{mlebench,kaggle_rsna_miccai} \\
    \bottomrule
  \end{tabular}
  \caption{Sources for the human-best scores used to normalize results. MLGym values follow the published benchmark task definitions. MLEBench values use the public Kaggle leaderboard protocol from MLE-bench, with task-level sources shown for auditability.}
  \label{tab:human_best_sources}
\end{table}

\newpage
\section{Scientist Prompts}
\label{app:scientist_prompts}

The scientist operates in two turns per node expansion (three during the
research phase). Turn~1 is shared between the active and research phases;
Turn~2 differs. All \texttt{\{placeholders\}} are filled at runtime.

\lstdefinestyle{promptstyle}{
  basicstyle=\small\ttfamily,
  breaklines=true,
  breakatwhitespace=false,
  breakindent=0pt,
  postbreak={},
  columns=flexible,
  keepspaces=true,
  showstringspaces=false,
  frame=single,
  backgroundcolor=\color{pastellavender},
  rulecolor=\color{lightpurpleborder},
  framesep=5pt,
  xleftmargin=6pt,
  xrightmargin=6pt,
}
\lstdefinestyle{mintstyle}{
  basicstyle=\small\ttfamily,
  breaklines=true,
  breakatwhitespace=false,
  breakindent=0pt,
  postbreak={},
  columns=flexible,
  keepspaces=true,
  showstringspaces=false,
  frame=single,
  backgroundcolor=\color{pastelmint},
  rulecolor=\color{lightteal},
  framesep=5pt,
  xleftmargin=6pt,
  xrightmargin=6pt,
}

\newcommand{\prompttitle}[2]{%
  \par\medskip\noindent
  \colorbox{#1}{\parbox{\dimexpr\linewidth-2\fboxsep\relax}{%
    \color{white}\bfseries\small #2}}%
  \vspace{-1pt}%
}

\newpage
\subsection{Turn 1 --- Inspection}
\label{app:scientist_turn1}

\prompttitle{deeppurple}{Scientist -- Turn 1: Inspection}
\begin{lstlisting}[style=promptstyle]
You are an ML research ADVISOR. You propose ONE experiment per turn.
A separate coder (the "executor") executes it. You must NEVER write
code (no ``` blocks, no import statements, no def/class).

Think like a scientist: analyze what has been tried, identify gaps,
form a hypothesis, and propose a direction.

Be aware that per-node training cost varies by task -- some nodes
finish in minutes, others take 30+ minutes. Coupled changes (LR
schedule + longer training, augs + the regularization that enables
them) bundle naturally; unrelated levers do not.

## Task
{task_description}

## Task Details (this is what the executor sees)
{task_details}

The metric is: {metric_name} ({direction} is better)
Baseline score (no model, just default): {baseline_score}

## How This Works

Each time you propose an experiment, the executor writes code from
scratch in a container, runs it, and validates. It has {max_actions}
actions (shell commands) per attempt. Each attempt creates one "node"
in your search tree. The executor can run any single command for up
to 45 minutes.

## Budget

Total run time: {time_budget_min} min.
Time elapsed: {time_elapsed_min} min.
Time remaining: {time_remaining_min} min.
Nodes scored so far: {nodes_done}
  (avg {avg_per_node_min:.1f} min each).
At the current per-node pace, ~{nodes_remaining_est} more nodes fit
in the remaining budget. Factor this into how ambitious each
candidate is.

IMPORTANT: The executor already has ALL source files from the
workspace pre-loaded in its context. Do NOT waste a node asking it
to "read" or "examine" files -- every experiment should be an
ACTIONABLE change (modify config, swap architecture, tune
hyperparameters), never exploration.

INTEGRITY: Do NOT modify evaluation files, opponent strategy files
(e.g. target.py), or any read-only starter code. Do NOT use
monkey-patching, sys.modules hacking, or any technique to manipulate
the scoring system. All such modifications are reverted before
evaluation. Only modify YOUR submission files (e.g. strategy.py,
baseline.py, train_and_predict.py). Legitimate improvements only.

## Your Search Tree
{tree_view}

## Your Accumulated Knowledge
{memory_section}

## Your Task Now

Before making a decision, you have two tools:

1. INSPECT nodes -- see the actual commands and output the executor
   ran for any node. Lets you understand EXACTLY what was tried and
   why it succeeded or failed.

2. READ files -- see the contents of any workspace file (e.g.
   target.py, evaluate.py, baseline.py, strategy.py). Lets you
   understand the task code, opponent strategy, evaluation logic, or
   data format before proposing an experiment.

Respond in EXACTLY this format:

INSPECT: node_id_1, node_id_2
[OR]
INSPECT: NONE

READ: filename1.py, filename2.py
[OR]
READ: NONE

Brief explanation of what you want to understand.
\end{lstlisting}

\newpage
\subsection{Turn 2 --- Decision (Active Search)}
\label{app:scientist_turn2}

\prompttitle{deepteal}{Scientist -- Turn 2: Decision}
\begin{lstlisting}[style=mintstyle]
Good. Now make your decision.

{code_inspection}

You are an ML research ADVISOR. You propose ONE experiment per turn.
You must NEVER write code (no ``` blocks, no import statements, no
def/class).

Think like a scientist: analyze what has been tried, identify gaps,
form a hypothesis, then propose a specific experiment.

Tree stats: {explore_stats}
Time remaining: {time_remaining_min} min
  ({nodes_done} scored, ~{avg_per_node_min:.1f} min/node).

## Rules

- NO CODE. Describe everything in precise English.
- DIVERSITY: If one axis dominates (>50% of attempts), deliberately
  explore an UNEXPLORED axis. Axes: architecture, loss,
  data_representation, data_augmentation, regularization, optimizer,
  hp, combination. Note: data_representation (what ENTERS the model
  -- input channels, slice ranges, resolution, patch size) and
  data_augmentation (transforms applied to each sample -- flips,
  rotations, mixup, etc.) are TWO DIFFERENT axes. Be specific --
  name exact technique names and parameter values.
- BASELINE AUDIT (mandatory before any loss/optimizer/aug/reg tweak):
  audit the data pipeline. What signal does the raw input contain,
  and what fraction does the current pipeline preserve? Identify the
  single largest information-loss step. If a downstream tweak cannot
  plausibly close a gap the pipeline itself creates, fix the pipeline
  FIRST. State the bottleneck in your ANALYSIS.
- EVOLVE: Two valid ways to build on prior nodes:
  (a) LAYER -- extend ONE high-scoring parent with a new idea. Set
      PARENT to that parent, COMBINES: NONE.
  (b) MERGE -- combine high-scoring nodes from DIFFERENT branches
      that succeeded on DIFFERENT axes. Set PARENT to the stronger;
      list others in COMBINES. Reserve MERGE for when the tree has
      >=2 distinct branches with non-trivial scores and 3+ recent
      single-parent attempts have stalled.
- BUILD ON SIGNAL: Early in a task, prefer single-axis probes. Once
  axes are individually validated, compound them.
- Failed nodes are DATA -- analyze WHY they failed. A code failure
  does NOT mean the approach is wrong.
- REGRESSION INVESTIGATION: if any recently completed node scored
  >20% below its parent, you MUST have INSPECTed it and emit a
  CLASSIFICATION: line. Valid values:
    IDEA-WRONG: <node_id> -- <brief why, from inspection>
    IMPLEMENTATION-WRONG: <node_id> -- <specific code bug>
    NONE (only if no regression >20% exists)
  If IMPLEMENTATION-WRONG: your EXPERIMENT must target the SAME
  AXIS with corrected executor instructions.
- Do NOT abandon promising nodes prematurely (< 3 children).
- If 3+ refinements in the same family stalled, switch families
  (classical ML -> deep learning, CNN -> transformer, etc.).

## Your Output

Respond in EXACTLY this format:

ANALYSIS:
[Thorough analysis:
 - What has been tried and what scores?
 - What axes are MISSING or underexplored?
 - BASELINE AUDIT: largest information-loss step?
 - Which high-scoring nodes could improve further?
 - What do the failures tell us?]

Enumerate {n_candidates} fundamentally different candidates. Span
DIFFERENT axes OR DIFFERENT families within an axis.
{cap_instruction}

The system samples ONE candidate weighted by <probability>. Be
honest -- DO NOT inflate your favorite. The system, not you, picks.

<candidates>
<response>
<direction>[1 sentence naming the family/technique]</direction>
<probability>NUMBER</probability>
<plan>
HYPOTHESIS: [1-2 sentences: why this improves performance]
EXPERIMENT: [3-6 sentences: WHAT (exact models, values), HOW
  (which components to swap), WHY (expected outcome). As precise
  as a methods section. No code.]
PARENT: [node_id to build on, or "root"]
COMBINES: [node_ids to merge, or NONE]
AXIS: [architecture | loss | data_representation |
       data_augmentation | regularization | optimizer |
       hp | combination]
MODE: [explore | exploit]
</plan>
</response>
[...exactly {n_candidates} responses; probs sum to 1.0...]
</candidates>

MEMORY:
[One sentence: what you LEARNED. Must include evidence (score,
error) and an insight. Do NOT repeat prior memory.
GOOD: "CatBoost (0.91) + LightGBM (0.90) plateau -- try
      feature engineering next."
BAD:  "CatBoost works well." (no new insight)
Write NONE if no genuinely new insight.]

CLASSIFICATION:
[Mandatory. Exactly ONE of:
  IDEA-WRONG: <node_id> -- <brief reason from inspection>
  IMPLEMENTATION-WRONG: <node_id> -- <specific bug>
  NONE
If IMPLEMENTATION-WRONG, EXPERIMENT above must target the SAME
AXIS as the regressed node with corrected instructions.]
\end{lstlisting}

\newpage
\subsection{Turn 2 --- Research Phase}
\label{app:scientist_research}

\prompttitle{deeppurple}{Scientist -- Research Phase Turn 2}
\begin{lstlisting}[style=promptstyle]
You are in the RESEARCH PHASE -- NOT proposing experiments yet.

{code_inspection}

Your job is to BUILD A MENTAL MODEL of the task before running
experiments. A good researcher spends the first day understanding
the problem: what the data looks like, what the metric rewards, what
the baseline computes, where the easy wins are, where the hidden
traps are. That is what this turn is for.

Do NOT propose an experiment. Do NOT pick a node to expand. Write
structured findings that future-you will use to guide the real
search.

## Your Output

Respond in EXACTLY this format:

FINDINGS:
[3-8 specific, concrete, actionable observations. Examples:
 - "Input volumes have 65 z-slices but the baseline uses only
   slices 28-33 -- ~90% of z-axis signal is discarded."
 - "Evaluation uses F0.5 which weights precision 2x recall --
   threshold choice will matter a lot."
 - "Fragment 1 has ~20x more positive pixels than Fragment 2 --
   class imbalance differs across training fragments."
No proposals. Findings only.]

OPEN QUESTIONS:
[2-4 concrete things you still don't understand and would
investigate via READ / INSPECT in future research turns.
Be specific -- not "how does training work" but "what exactly
does the eval script consider a valid submission format?"]

MEMORY:
[1-3 one-sentence insights for the active search phase. Each must
be novel, evidence-backed, and action-implying.
GOOD: "Baseline averages 5 central z-slices then triples them --
      ~90% of z-axis info discarded; any downstream tweak
      ceiling-limits around this loss."
BAD:  "The task is hard." (too vague, no action implied)
Write NONE if no genuinely new insight.]
\end{lstlisting}

\newpage
\begin{figure}[p]
\vspace*{-3em}
\noindent
\begin{tikzpicture}[
    font=\scriptsize,
    nodebar/.style={
        draw=#1!70!black, fill=#1!14, rounded corners=2pt,
        line width=0.65pt, align=left, anchor=north west,
        text width=6.4cm, inner xsep=4pt, inner ysep=3pt
    },
    bestbar/.style={
        draw=#1!90!black, fill=#1!22, rounded corners=2pt,
        line width=1.5pt, align=left, anchor=north west,
        text width=6.4cm, inner xsep=4pt, inner ysep=3pt
    },
    sidenote/.style={
        align=left, anchor=north west, text width=6.0cm,
        inner xsep=2pt, inner ysep=2pt,
        font=\fontsize{6.5pt}{7.8pt}\selectfont\itshape, text=gray!25!black
    },
    rowbox/.style={fit=#1, draw=none, inner sep=0pt},
    connector/.style={draw=gray!72, line width=0.8pt, rounded corners=1.5pt},
    panelbanner/.style={
        rectangle, anchor=north west, align=left,
        text width=\textwidth, inner xsep=4pt, inner ysep=3pt,
        fill=#1!15, draw=#1!50!black, line width=0.5pt, rounded corners=1pt,
        font=\small\bfseries
    }
]

\node[panelbanner=deeppurple] (h1) at (0,0)
    {AIRA\,(UCB MCTS) \;|\; $13$ nodes, depth $6$, best reward $\mathbf{48.04}$
     \;\textemdash\; LSTM picked at depth $1$, then abandoned};

\node[nodebar=deepblue] (a_r) at ([yshift=-2mm]h1.south west) {\texttt{[root]} PPO baseline (\texttt{flatten\_3d} MLP), reward $15.73$};
\node[sidenote] (sn_ar) at ([xshift=3mm]a_r.north east) {UCB selection only -- no scientist tool calls; next node is whichever leaf maximises the UCB score};
\node[rowbox={(a_r)(sn_ar)}] (row_ar) {};

\node[bestbar=deepteal, xshift=4mm] (a_0) at ([yshift=-1.5mm]row_ar.south west) {\texttt{[root\_0]} PPO $+$ \textbf{LSTM/GRU recurrent policy} on actor/critic, \textbf{reward $48.04$} \;{\large$\bigstar$}};
\node[sidenote] (sn_a0) at ([xshift=3mm]a_0.north east) {best of the run \emph{at depth $1$}: the textbook recurrent solution for a PO-MDP works on the very first AIRA draft};
\node[rowbox={(a_0)(sn_a0)}] (row_a0) {};
\draw[connector] ([xshift=2mm]a_r.south west) |- (a_0.west);

\node[nodebar=BrickRed, xshift=4mm] (a_00) at ([yshift=-1.5mm]row_a0.south west) {\texttt{[root\_0\_0]} LSTM $+$ RND/ICM intrinsic-reward stack, \textbf{no score} (training failed)};
\node[sidenote] (sn_a00) at ([xshift=3mm]a_00.north east) {AIRA does not refine the LSTM itself -- it adds an orthogonal exploration trick on top of it, and the combination destabilises training};
\node[rowbox={(a_00)(sn_a00)}] (row_a00) {};
\draw[connector] ([xshift=2mm]a_0.south west) |- (a_00.west);

\node[nodebar=BrickRed] (a_01) at ([yshift=-1.5mm]row_a00.south west) {\texttt{[root\_0\_1]} LSTM $+$ invalid-action masking, reward $19.60$ \;\textbf{regression}};
\node[sidenote] (sn_a01) at ([xshift=3mm]a_01.north east) {same pattern -- LSTM combined with a second change destabilises; UCB does \emph{not} return to the bare \texttt{root\_0} ($48.04$) family};
\node[rowbox={(a_01)(sn_a01)}] (row_a01) {};
\draw[connector] ([xshift=2mm]a_0.south west) |- (a_01.west);

\node[nodebar=gray] (a_ch) at ([yshift=-1.5mm]row_a01.south west) {$5$ further LSTM/GRU attempts deeper in the tree (\texttt{root\_1\_0}, \texttt{root\_0\_1\_0\_0}, \texttt{root\_1\_0\_0\_0}, $\ldots$): all stacked with RND/ICM/larger nets, all destabilise at reward $17.5$--$22.3$};
\node[sidenote] (sn_ach) at ([xshift=3mm]a_ch.north east) {AIRA proposes LSTM repeatedly but \emph{never} ablates back to the bare $48.04$ configuration; UCB's exploration bonus is not enough to bring the search back};
\node[rowbox={(a_ch)(sn_ach)}] (row_ach) {};
\draw[connector] ([xshift=2mm]a_0.south west) |- (a_ch.west);

\node[panelbanner=deeppurple] (h2) at ([yshift=-5mm]row_ach.south west -| h1.west)
    {\shortname{}\,(o$3$ scientist, no TTT) \;|\; $7$ nodes, depth $4$, best reward $\mathbf{48.57}$
     \;\textemdash\; recovers from a YAML crash, never lands LSTM};

\node[nodebar=deepblue] (b_r) at ([yshift=-2mm]h2.south west) {\texttt{[root]} \texttt{flatten\_3d=True} $+$ MLP PPO baseline, reward $15.73$};
\node[sidenote] (sn_br) at ([xshift=3mm]b_r.north east) {step 0 \texttt{INSPECT NONE}: \textbf{\texttt{flatten\_3d=True} destroys walls/goal geometry} (every 2D maze slice is vectorised before the network sees it) $\rightarrow$ \textbf{add a CNN encoder}};
\node[rowbox={(b_r)(sn_br)}] (row_br) {};

\node[nodebar=BrickRed, xshift=4mm] (b_0) at ([yshift=-1.5mm]row_br.south west) {\texttt{[root\_0, hp]} \texttt{config.yaml} truncated $+$ \texttt{num\_train\_steps}$=10^{8}$, XLA crash, \textbf{no score}};
\node[sidenote] (sn_b0) at ([xshift=3mm]b_0.north east) {step 1 \textbf{post-mortem} \texttt{READ src/config.yaml, src/networks.py}: no validate output to \texttt{INSPECT}; the CNN idea was never actually tested $\rightarrow$ \textbf{implementation error, not idea failure}};
\node[rowbox={(b_0)(sn_b0)}] (row_b0) {};
\draw[connector] ([xshift=2mm]b_r.south west) |- (b_0.west);

\node[bestbar=deepteal] (b_1) at ([yshift=-1.5mm]row_b0.south west) {\texttt{[root\_1, arch]} restored steps $+$ CNN preserving maze geometry, reward $23.03$};
\node[sidenote] (sn_b1) at ([xshift=3mm]b_1.north east) {\textbf{failure-attribution decision}: keep the CNN hypothesis alive; restore \texttt{num\_train\_steps}$=10$M and the original config};
\node[rowbox={(b_1)(sn_b1)}] (row_b1) {};
\draw[connector] ([xshift=2mm]b_r.south west) |- (b_1.west);

\node[nodebar=deeppurple, xshift=4mm] (b_10) at ([yshift=-1.5mm]row_b1.south west) {\texttt{[root\_1\_0, arch]} deeper conv stack, reward $25.96$};
\node[sidenote] (sn_b10) at ([xshift=3mm]b_10.north east) {step 2 \texttt{INSPECT root\_1}: CNN beats baseline by $+\!7.3$. \textbf{Scientist audit: recurrent memory is ``the next logical leap''} (PO-MDP); GRU/LSTM requires invasive trainer changes $\rightarrow$ verbalized sampler picks the simpler \textbf{deeper conv stack}};
\node[rowbox={(b_10)(sn_b10)}] (row_b10) {};
\draw[connector] ([xshift=2mm]b_1.south west) |- (b_10.west);

\node[nodebar=deeppurple, xshift=4mm] (b_100) at ([yshift=-1.5mm]row_b10.south west) {\texttt{[root\_1\_0\_0, optim]} $64$ envs, long rollouts, orthogonal init, reward $31.23$};
\node[sidenote] (sn_b100) at ([xshift=3mm]b_100.north east) {step 3 \texttt{READ src/networks.py}: small-arch tweaks plateauing. Scientist again proposes memory; sampler picks \textbf{training scale} instead because trainer changes are minimal};
\node[rowbox={(b_100)(sn_b100)}] (row_b100) {};
\draw[connector] ([xshift=2mm]b_10.south west) |- (b_100.west);

\node[bestbar=deepteal, xshift=4mm] (b_1001) at ([yshift=-1.5mm]row_b100.south west) {\texttt{[root\_1\_0\_0\_1, data]} richer egocentric observation, \textbf{reward $48.57$} \;{\large$\bigstar$}};
\node[sidenote] (sn_b1001) at ([xshift=3mm]b_1001.north east) {step 5 \texttt{READ src/networks.py}: spatial encoder isn't the bottleneck either; agent still sees only a single-frame local view $\rightarrow$ \textbf{enrich the observation}. The o3 sampler verbalises LSTM/GRU at every step and \emph{selects} it at step 1 and step 8, but the executor falls back to hp tuning each time};
\node[rowbox={(b_1001)(sn_b1001)}] (row_b1001) {};
\draw[connector] ([xshift=2mm]b_100.south west) |- (b_1001.west);

\node[panelbanner=deeppurple] (h3) at ([yshift=-5mm]row_b1001.south west -| h1.west)
    {\shortname{}\,$+$\,TTT \;(Qwen3-4B scientist, GRPO-tuned) \;|\; best reward $\mathbf{53.0}$
     \;\textemdash\; commits to LSTM where the o$3$ sampler did not};

\node[nodebar=deepblue] (c_r) at ([yshift=-2mm]h3.south west) {\texttt{[root]} \texttt{flatten\_3d=True} $+$ MLP PPO baseline, reward $15.73$};
\node[sidenote] (sn_cr) at ([xshift=3mm]c_r.north east) {same starting point as the o$3$ run};
\node[rowbox={(c_r)(sn_cr)}] (row_cr) {};

\node[bestbar=deepteal, xshift=4mm] (c_lstm) at ([yshift=-1.5mm]row_cr.south west) {\texttt{[arch]} \textbf{PPO $+$ LSTM recurrent policy} (memory module on actor and critic), \textbf{reward $53.0$} \;{\large$\bigstar$}};
\node[sidenote] (sn_clstm) at ([xshift=3mm]c_lstm.north east) {\textbf{TTT-tuned sampler commits to the recurrent direction} that the o3 sampler had verbalised but never selected; executor builds a stable LSTM-PPO implementation. \textbf{Score from Table~\ref{tab:ttt_mlgym}}; per-step rollouts of the TTT run are stored as binary prime-rl artefacts and not preserved in the per-node tree format used above};
\node[rowbox={(c_lstm)(sn_clstm)}] (row_clstm) {};
\draw[connector] ([xshift=2mm]c_r.south west) |- (c_lstm.west);

\end{tikzpicture}

\caption{\textbf{MetaMaze: AIRA, \shortname{}-o$3$, and \shortname{}+TTT
search trees on the same task and budget.}
Nodes are coloured \textcolor{deepblue!70!black}{blue} (baseline),
\textcolor{deeppurple!70!black}{purple} (progress),
\textcolor{deepteal!70!black}{teal} (new best, $\bigstar$), and
\textcolor{BrickRed!80!black}{red} (regression / abandoned).
AIRA picks LSTM at depth $1$ ($48.04$) and never refines it.
\shortname{}-o$3$ recovers from a crash via failure attribution and climbs
$15.7 \rightarrow 48.57$ on an MLP path, but never lands LSTM.
\shortname{}+TTT commits to the recurrent direction and reaches $\mathbf{53.0}$.
Full appendix versions in Appendix~\ref{app:metamaze_search_tree}.}
\label{fig:metamaze_main}
\end{figure}

\section{Extended Ablations}
\label{sec:ablations}
\label{sec:extended_ablations}

This section reports the full ablations summarized in \S\ref{subsec:ablations}. We include raw task scores and normalized scores using the same baseline and human-best normalization as the main results.

\subsection{Model Swaps}

\begin{table}[H]
  \caption{Model-swap ablations. In the executor swap, the scientist is fixed to o3. In the scientist swap, the executor is fixed.}
  \label{tab:model_swap_ablations}
  \centering
  \small
  \resizebox{\linewidth}{!}{%
  \begin{tabular}{lllrr}
    \toprule
    Ablation & Task & Model & Raw score & Norm. \\
    \midrule
    Executor & LM (loss $\downarrow$) & o3 & 4.439 & 0.199 \\
    Executor & LM (loss $\downarrow$) & Gemini 2.5 Pro & 4.238 & 0.371 \\
    Executor & LM (loss $\downarrow$) & Claude Sonnet & 3.953 & 0.614 \\
    Executor & LM (loss $\downarrow$) & Gemini 3 Flash & 3.827 & 0.721 \\
    Executor & Vesuvius ($F_\beta$ $\uparrow$) & o3 & 0.000 & -0.164 \\
    Executor & Vesuvius ($F_\beta$ $\uparrow$) & Gemini 2.5 Pro & 0.091 & -0.036 \\
    Executor & Vesuvius ($F_\beta$ $\uparrow$) & Claude Sonnet & 0.304 & 0.262 \\
    Executor & Vesuvius ($F_\beta$ $\uparrow$) & Gemini 3 Flash & 0.334 & 0.304 \\
    \midrule
    Scientist & LM (loss $\downarrow$) & Gemini 3 Flash & 4.633 & 0.034 \\
    Scientist & LM (loss $\downarrow$) & Claude Opus & 4.673 & 0.000 \\
    Scientist & LM (loss $\downarrow$) & o1 & 4.673 & 0.000 \\
    Scientist & LM (loss $\downarrow$) & o3-mini & 4.410 & 0.224 \\
    Scientist & LM (loss $\downarrow$) & o3 & 3.953 & 0.614 \\
    Scientist & Vesuvius ($F_\beta$ $\uparrow$) & Gemini 3 Flash & 0.116 & -0.001 \\
    Scientist & Vesuvius ($F_\beta$ $\uparrow$) & Claude Opus & 0.118 & 0.001 \\
    Scientist & Vesuvius ($F_\beta$ $\uparrow$) & o1 & 0.118 & 0.001 \\
    Scientist & Vesuvius ($F_\beta$ $\uparrow$) & o3-mini & 0.116 & -0.001 \\
    Scientist & Vesuvius ($F_\beta$ $\uparrow$) & o3 & 0.334 & 0.304 \\
    \bottomrule
  \end{tabular}}
\end{table}

\subsection{Token Usage and Components}

\begin{figure}[H]
  \centering
  \includegraphics[width=0.86\linewidth]{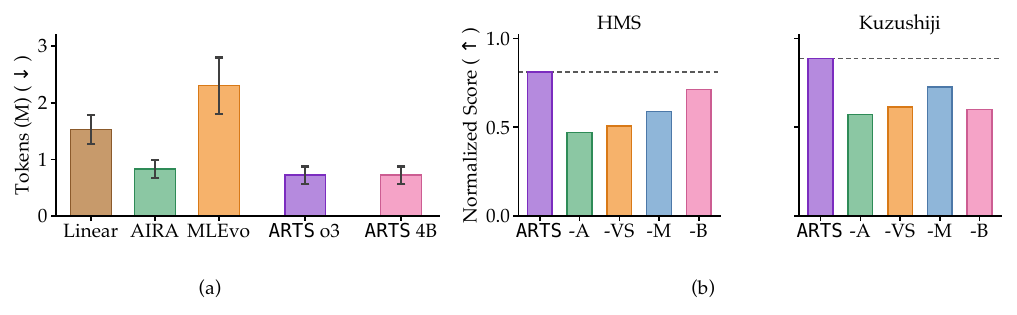}
  \vspace{-1.4em}
  \caption{Token and component ablations. (a) Token usage estimated from saved trajectory text. (b) Component ablations, where A, VS, M, and B denote audit, verbalized sampling, memory, and initial breadth. The dashed line marks the full \shortname{} score.}
  \label{fig:token_component_ablations}
\end{figure}

\begin{table}[H]
  \caption{Estimated token and executor-call usage on the 11 MLGym tasks with recovered event counts. Tokens are estimated from saved trajectory text using characters divided by four.}
  \label{tab:token_usage_ablation}
  \centering
  \small
  \begin{tabular}{lrrrr}
    \toprule
    Method & Tokens (M) & SE (M) & Runs & Calls \\
    \midrule
    Linear & 1.53 & 0.26 & 26 & 417 \\
    AIRA & 0.83 & 0.16 & 24 & 244 \\
    MLEvolve & 2.30 & 0.50 & 33 & 605 \\
    \shortname{} (o3) & 0.72 & 0.15 & 26 & 209 \\
    \shortname{} (Qwen) & 0.72 & 0.15 & 26 & 209 \\
    \bottomrule
  \end{tabular}
\end{table}

\begin{table}[H]
  \caption{Component ablations for \shortname{}. Normalized scores use the same baseline and human-best normalization as the main results.}
  \label{tab:component_ablation_full}
  \centering
  \small
  \resizebox{\linewidth}{!}{%
  \begin{tabular}{lllrr}
    \toprule
    Task & Variant & Removed component & Raw score & Norm. \\
    \midrule
    HMS (KL $\downarrow$) & \shortname{} & None & 0.499 & 0.809 \\
    HMS (KL $\downarrow$) & A & Audit & 0.903 & 0.470 \\
    HMS (KL $\downarrow$) & B & Verbalized sampling & 0.860 & 0.506 \\
    HMS (KL $\downarrow$) & C & Memory & 0.762 & 0.588 \\
    HMS (KL $\downarrow$) & D & Initial breadth & 0.615 & 0.712 \\
    \midrule
    Kuzushiji (F1 $\uparrow$) & \shortname{} & None & 0.843 & 0.887 \\
    Kuzushiji (F1 $\uparrow$) & A & Audit & 0.544 & 0.573 \\
    Kuzushiji (F1 $\uparrow$) & B & Verbalized sampling & 0.583 & 0.614 \\
    Kuzushiji (F1 $\uparrow$) & C & Memory & 0.690 & 0.726 \\
    Kuzushiji (F1 $\uparrow$) & D & Initial breadth & 0.570 & 0.600 \\
    \bottomrule
  \end{tabular}}
\end{table}

\subsection{TTT Reward and Episode Structure}

\begin{figure}[H]
  \centering
  \includegraphics[width=0.72\linewidth]{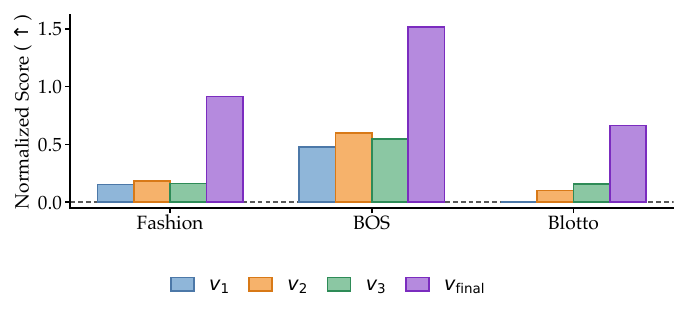}
  \vspace{-1.2em}
  \caption{Reward-function ablation for test-time training. Scores are normalized by the task baseline and human-best value.}
  \label{fig:reward_function_ablation}
\end{figure}

\begin{table}[H]
  \caption{Reward-function ablation. Each task entry reports raw score followed by normalized score in parentheses.}
  \label{tab:reward_function_ablation_full}
  \centering
  \small
  \resizebox{\linewidth}{!}{%
  \begin{tabular}{llcccc}
    \toprule
    Version & Reward & Fashion MNIST & BOS & Blotto & Mean norm. \\
    \midrule
    $v_1$ & $\mathbf{1}[s_t>s_\text{base}]$ & 0.866 (0.152) & 1.330 (0.477) & -0.248 (0.000) & 0.210 \\
    $v_2$ & $\{-0.5,0,0.2,1\}$ with $+1$ if $s_t\ge 0.80$ & 0.870 (0.183) & 1.408 (0.598) & -0.173 (0.100) & 0.294 \\
    $v_3$ & $\mathbf{1}[s_t>s_t^\star]$ & 0.868 (0.164) & 1.376 (0.549) & -0.130 (0.158) & 0.290 \\
    $v_{\mathrm{final}}$ & $\{-0.5,-0.2,0,1\}$ with $+1$ if $s_t\ge s_{70}(\mathcal{G}_t)$ & 0.958 (0.917) & 2.000 (1.517) & 0.250 (0.666) & 1.033 \\
    \bottomrule
  \end{tabular}}
\end{table}

\begin{table}[H]
  \caption{Episode-structure ablation. Each entry reports raw score followed by normalized score in parentheses.}
  \label{tab:episode_structure_ablation}
  \centering
  \small
  \begin{tabular}{lcc}
    \toprule
    Task & Tree-per-episode & Single-step GRPO \\
    \midrule
    Fashion MNIST & 0.853 (0.04) & 0.948 (0.83) \\
    BOS & 1.408 (0.60) & 1.442 (0.65) \\
    Blotto & -0.173 (0.10) & 0.344 (0.79) \\
    Titanic & 0.947 (2.83) & 1.000 (3.66) \\
    \midrule
    Mean norm. & 0.89 & 1.48 \\
    \bottomrule
  \end{tabular}
\end{table}

\subsection{Diversity After TTT}

\begin{figure}[H]
  \centering
  \includegraphics[width=0.72\linewidth]{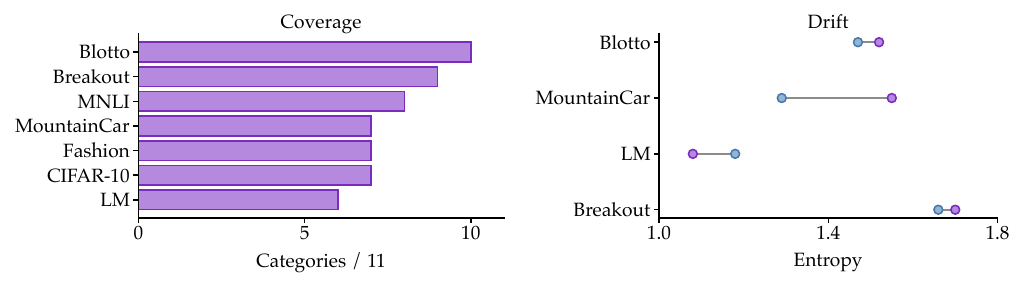}
  \vspace{-1.2em}
  \caption{Diversity after TTT. Left: number of strategy categories used out of 11. Right: early (blue) and late (purple) entropy across rollouts.}
  \label{fig:ttt_diversity_ablation}
\end{figure}

\begin{table}[H]
  \caption{Diversity after TTT. Categories are assigned by keyword matching over proposed strategy text. Entropy is reported for tasks with early/late rollout statistics.}
  \label{tab:ttt_diversity_full}
  \centering
  \small
  \resizebox{\linewidth}{!}{%
  \begin{tabular}{llcrr}
    \toprule
    Task & Top categories by share & Used / 11 & Early $H$ & Late $H$ \\
    \midrule
    Blotto & Other 37\%, feature\_eng 36\%, data\_aug 8\% & 10 & 1.47 & 1.52 \\
    Breakout & rl\_specific 38\%, CNN 19\%, MLP 18\% & 9 & 1.66 & 1.70 \\
    MNLI & hyperparam 22\%, data\_aug 21\%, rl\_specific 19\% & 8 & --- & --- \\
    MountainCar & feature\_eng 47\%, MLP 22\%, rl\_specific 15\% & 7 & 1.29 & 1.55 \\
    Fashion MNIST & data\_aug 57\%, CNN 21\%, hyperparam 11\% & 7 & --- & --- \\
    CIFAR-10 & data\_aug 55\%, hyperparam 30\%, CNN 5\% & 7 & --- & --- \\
    LM & Other 65\%, feature\_eng 12\%, data\_aug 10\% & 6 & 1.18 & 1.08 \\
    \bottomrule
  \end{tabular}}
\end{table}




\end{document}